%% file: acl_latex.tex
\pdfoutput=1

\documentclass[11pt]{article}

\usepackage[final]{acl}

\usepackage{times}
\usepackage{latexsym}

\usepackage[T1]{fontenc}

\usepackage[utf8]{inputenc}

\usepackage{microtype}

\usepackage{inconsolata}

\usepackage{graphicx}

\usepackage{bm}
\usepackage{amssymb}
\usepackage{amsmath}
\usepackage{pifont}
\usepackage{tikz}
\usetikzlibrary{shapes.geometric}
\usetikzlibrary{arrows.meta}
\usetikzlibrary{calc, positioning}
\usepackage{pgfplots}
\usepgfplotslibrary{groupplots}
\usepackage{subcaption}
\usepackage{enumitem}
\usepackage{multirow}
\usepackage{algorithm}
\usepackage{algcompatible}
\usepackage{booktabs}

\title{MobiZO: Enabling Efficient LLM Fine-Tuning at the Edge \\ via Inference Engines}

\author{Lei Gao $^\ast$,  Amir Ziashahabi\thanks{These authors contributed equally.}, Yue Niu, Salman Avestimehr,  Murali Annavaram \\
        University of Southern California\\ 
        \texttt{\{leig, ziashaha, yueniu, avestime, annavara\}@usc.edu}
        }

\begin{document}
\maketitle
\begin{abstract}
Large Language Models (LLMs) are currently pre-trained and fine-tuned on large cloud servers. The next frontier is LLM personalization, where a foundation model can be fine-tuned with user/task-specific data. Given the sensitive nature of such private data, it is desirable to fine-tune these models on edge devices to improve user trust. However, fine-tuning on resource-constrained edge devices presents significant challenges due to substantial memory and computational demands, as well as limited infrastructure support. We observe that inference engines (e.g., ExecuTorch) can be repurposed for fine-tuning by leveraging zeroth-order (ZO) optimization, which uses multiple forward passes to approximate gradients. While promising, direct application of ZO methods on edge devices is inefficient due to the high computational cost of multiple forward passes required for accurate gradient estimation, and their deployment has been largely unexplored in practice. We introduce MobiZO, a resource-efficient fine-tuning framework for LLMs specifically designed for edge devices. 
MobiZO combines three key innovations: (1) a parallelized randomized gradient estimator that employs both outer-loop and inner-loop parallelism to eliminate sequential forward passes, (2) a specialized Multi-Perturbed LoRA (MP-LoRA) module that enables efficient realization of both inner and outer loop parallelism, and (3) a seamless integration with ExecuTorch for on-device training, requiring no modifications to the runtime.
Experiments demonstrate that MobiZO achieves substantial runtime speedups and memory savings while improving fine-tuning accuracy, paving the way for practical deployment of LLMs in real-time, on-device applications. Code available at: \url{https://github.com/leigao97/MobiZO}.

\end{abstract}

\section{Introduction} \label{sec:intro}

Large Language Models (LLMs) have demonstrated strong performance across varied tasks, chatbots, image generation \cite{openai2024gpt4technicalreport, chowdhery2022palm, geminiteam2024gemini}. Fine-tuning is a crucial step for adapting LLMs to specific tasks, but it demands significant memory resources for storing model parameters, gradients, activations, and optimizer states \cite{wan2024efficient}. This memory overhead makes fine-tuning infeasible on resource-constrained devices such as smartphones and edge platforms \cite{yin2024llmservicemobiledevices}. Moreover, existing on-device frameworks like ExecuTorch \cite{executorch} and TensorFlow Lite \cite{tensorflow2020} primarily optimize inference, leaving fine-tuning largely unsupported.

\textbf{Resource Challenges Despite Recent Advances.} Techniques such as parameter-efficient fine-tuning (PEFT) \cite{hu2022lora, houlsby2019adapter, li2021prefix, lester2021ptuning} and memory-efficient fine-tuning \cite{dettmers2023qlora, lv2023full, zhao2024galore, malladi2023finetuning} can significantly reduce the memory footprint associated with model weights, gradients, and optimizer states. However, even with these methods, storing internal activations during backpropagation remains a significant challenge. For example, fine-tuning Llama 7B requires up to 45.6~GB of on-chip memory for internal activations \cite{lv2023full}, making it impractical for most edge devices. Current solutions still fall short of meeting the stringent resource constraints of edge environments.

\textbf{Limitations of On-Device Training Frameworks.} While several techniques exist to mitigate the memory costs of intermediate activations during backpropagation, they generally rely on training frameworks that support automatic differentiation to perform backpropagation. For instance, gradient checkpointing \cite{chen2016trainingdeepnetssublinear} discards select activations during the forward pass and recomputes them during backpropagation, while gradient accumulation aggregates gradients over smaller batches. PockEngine \cite{zhu2023pockengine} limits backpropagation to update a subset of layers, reducing the need to store activations for other layers. However, all these techniques are not well supported by the existing on-device training frameworks on most edge platforms such as Android devices.

\textbf{Zeroth-Order Optimization as a Potential Solution.} Zeroth-order (ZO) optimization has gained attention as a way to eliminate the need to store activations by estimating gradients using only forward passes. Specifically, ZO methods approximate gradients by evaluating the loss function at multiple perturbed versions of the model weights and using these values for gradient estimation. This approach has the potential to solve the memory challenge, as well as avoid the need for backpropagation support. Thus, ZO methods hold promise for on-device fine-tuning by utilizing existing inference frameworks like ExecuTorch \cite{executorch}. However, applying ZO optimization to fine-tune LLMs on edge devices presents its own set of challenges.

One classic zeroth-order optimizer, the Randomized Gradient Estimator (RGE) \cite{duchi2015, nesterov2017}, estimates gradients by computing finite differences of function values along randomly chosen perturbation vectors. With RGE, estimation accuracy for each step of training improves as the number of stochastic perturbations (also referred to as queries) increases \cite{zhang2024revisiting, gautam2024variancereduced, yang2024adazeta}, but the computational cost scales linearly with the query count. In addition, its on-device adaptation remains largely unexplored.

In this work, we propose \textbf{MobiZO} training framework to address the runtime overhead inherent in multi-query RGE. MobiZO includes a novel \textbf{Multi-Perturbed LoRA} (MP-LoRA) design and combines \textbf{outer-loop parallelization} and \textbf{inner-loop parallelization} to perform multiple forward passes in parallel, substantially reducing per-step latency while harvesting the accuracy benefits of multi-query gradient estimation. Moreover, MobiZO allows seamless adaptation for deploying RGE optimization via inference engines to enable practical on-device fine-tuning.
Our contributions are as follows:

\begin{itemize}[leftmargin=*] 
\item We introduce the MobiZO framework, specialized for on-device training, consisting of outer-loop, inner-loop parallelization, and Multi-Perturbed LoRA designs. By executing multiple forward passes in parallel within each MP-LoRA module, MobiZO effectively amortizes the memory access cost of loading model parameters, thereby reducing training time while improving model performance.

\item We demonstrate that the MobiZO framework can be seamlessly integrated into inference engines such as ExecuTorch without requiring any modifications to its runtime code. Our approach is realized through minimal server-side code changes only, making it practical for on-device fine-tuning.

\item We empirically validate that our method achieves substantial wall-clock time speedups and memory savings while improving model performance. Our approach results in up to \textbf{4.3$\times$} end-to-end training speedups and up to \textbf{8.1}$\%$ improvement in accuracy compared to the MeZO baseline.
\end{itemize}

\section{Background and Related Work}
\label{sec:background}

\textbf{Low-Rank Adaptation.} To reduce the resource demands of LLM fine-tuning, parameter-efficient fine-tuning methods update only a small subset of parameters. LoRA \cite{hu2022lora} introduces trainable low-rank matrices $\mathbf{A} \in \mathbb{R}^{k_{in} \times r}$ and $\mathbf{B} \in \mathbb{R}^{r \times k_{out}}$ while freezing the original weight matrix $\mathbf{W}$. Since $r \ll \min(k_{in}, k_{out})$, the number of trainable parameters is significantly reduced. The forward pass is computed as $\mathbf{y} = \mathbf{x}\mathbf{W} + \mathbf{x}\mathbf{A}\mathbf{B}$, where $\mathbf{A}$ is initialized randomly and $\mathbf{B}$ starts at zero, ensuring no initial deviation from the pre-trained model. Variations such as LoRA-FA \cite{zhang2023lorafa} further reduce trainable parameters by freezing $\mathbf{A}$ and updating only $\mathbf{B}$.

\textbf{Zeroth-Order Optimization.} ZO optimization methods have been widely applied across various machine learning applications \cite{chen2017zoo, sun2022bbtv2, wang2022zarts, liu2024dpzo}. Among ZO gradient estimators, the randomized gradient estimator (RGE) is particularly effective, especially for fine-tuning LLMs \cite{malladi2023finetuning}.
Given a labeled dataset $\mathcal{D}$ and a model with parameters $\bm{\theta} \in \mathbb{R}^d$, let the loss function on a minibatch $\bm{\mathcal{B}} \subset \mathcal{D}$ of size $B$ be denoted as $\mathcal{L}(\bm{\theta}; \bm{\mathcal{B}})$. The RGE estimates the gradient of the loss $\mathcal{L}$ with respect to the parameters $\bm{\theta}$ on a minibatch $\bm{\mathcal{B}}$ via:

\small
\begin{equation*}
\hat{\nabla} \mathcal{L}(\bm{\theta}; \bm{\mathcal{B}}) = \frac{1}{q} \sum_{i=1}^q \left[ \frac{\mathcal{L}(\bm{\theta} + \epsilon \bm{z}_i; \bm{\mathcal{B}}) - \mathcal{L}(\bm{\theta} - \epsilon \bm{z}_i; \bm{\mathcal{B}})}{2 \epsilon} \bm{z}_i \right],
\end{equation*}
\normalsize
where $\bm{z}_i \sim \mathcal{N}(0, \bm{I}{d})$, $q$ is the query number, and $\epsilon$ is the perturbation scale. 
The choice of $q$ balances the variance of the ZO gradient estimate and the computational cost, and the variance of the RGE is approximately $O(d/q)$ \cite{zhang2024revisiting}.

ZO-SGD replaces FO gradients with ZO gradient estimates: $\bm{\theta}_{t+1} = \bm{\theta}_{t} - \eta \hat{\nabla} \mathcal{L}(\bm{\theta}; \bm{\mathcal{B}}_{t})$, with learning rate $\eta$ at timestep $t$.
The choice of optimizer (SGD) is orthogonal to ZO optimization methods, but in our preliminary experiments, we find adaptive optimizers such as Adam would not necessarily improve LLM fine-tuning performance.

\textbf{ZO LLM Fine-Tuning.} Conventional RGE training requires storing perturbation noise $\bm{z}$, effectively doubling inference memory. MeZO \cite{malladi2023finetuning} eliminates this overhead by storing only the random seed and regenerating $\bm{z}$ on demand. While MeZO also considers $q>1$, it compensates for the increased computation per step by proportionally reducing the total number of training steps (e.g., halving the steps when $q=2$). Under this fixed computational budget, they observe that larger $q$ does not improve accuracy compared to $q=1$, prompting MeZO to adopt $q=1$ as the default setting. In contrast, \citeauthor{zhang2024revisiting} benchmarked various ZO optimization methods, including RGE with $q>1$, and confirmed that when computational constraints are lifted, larger $q$ can indeed enhance performance.

Sparse-MeZO \cite{liu2024sparsemezo} selectively updates parameters but is sensitive to hyperparameters. Extreme-sparse-MeZO \cite{guo2024extremesparse} integrates first-order Fisher-based sparse training. MeZO-SVRG \cite{gautam2024variancereduced} improves variance reduction but occasionally requires full-dataset gradient estimation, increasing cost. AdaZeta \cite{yang2024adazeta} adaptively schedules queries but still relies on sequential gradient estimations.

\textbf{On-device LLM Training.} Several methods address the memory and compute constraints of on-device LLM training. PockEngine \cite{zhu2023pockengine} updates only select layers, skipping gradient calculations for less critical parameters. FwdLLM \cite{xu2024fwdllm} applies numerical differentiation to approximate gradients, lowering communication costs in federated learning, but is limited to CUDA environment. HETLORA \cite{cho2024hetlora} enables federated LoRA training across heterogeneous devices but requires further real-world testing due to high activation memory costs. PocketLLM \cite{peng2024pocketllm} evaluates MeZO for on-device fine-tuning but does so in a simulated Linux environment rather than on mobile devices.

\begin{algorithm}[t]
\caption{MobiZO Algorithm.}
\label{alg:mobizo}
\begin{algorithmic}[1]
\STATE {\bfseries Input:} \ learnable parameters $\bm{\theta}_l \in \mathbb{R}^{d_l}$, frozen parameters $\bm{\theta}_f \in \mathbb{R}^{d_f}$, loss $\mathcal{L}: \mathbb{R}^{d_l} \times \mathbb{R}^{d_f} \rightarrow \mathbb{R}$, step budget $T$, query budget $q$, effective batch size $E$, perturbation scale $\epsilon$, learning rate $\eta$
\FOR{$t = 1$ to $T$}
    \STATE Sample batch $\bm{\mathcal{B}} \subset \mathcal{D}$
    \FOR {$i = 1$ to $q$}{\space \textbf{in parallel} {\color{blue} $\triangleright$ Outer}}
        \STATE Sample random seed $s_i$ 
        \STATE $\bm{z}_i {\sim} \mathcal{N}(0, \bm{I}_{d_l})$ using $s_i$
        \FOR {$k \in \{+1, -1\}$}{\space \textbf{in parallel} {\color{blue} $\triangleright$ Inner}}
            \STATE $\bm{\theta}_l^{(k)} = \bm{\theta}_l + k \epsilon \bm{z}_i$ {\color{blue} $\triangleright$ MP-LoRA}
            \STATE $\ell^{(k)} = \mathcal{L}((\bm{\theta}_l^{(k)}, \bm{\theta}_f); \bm{\mathcal{B}} )$
        \ENDFOR
        \STATE $g_i = \dfrac{\ell^{(+1)} - \ell^{(-1)}}{2\epsilon}$ 
        \STATE Store $s_i$ and $g_i$
    \ENDFOR
    \STATE $\bm{\theta}_l \gets \bm{\theta}_l - \eta \left( \dfrac{1}{q} \sum\limits_{i=1}^q g_i \bm{z}_i \right)$
\ENDFOR
\end{algorithmic}
\end{algorithm}

\section{The MobiZO Framework}
\label{sec:method}

To perform a $q$-query gradient estimation, Randomized Gradient Estimation (RGE) typically requires $2q$ forward passes. The naive implementation of RGE (detailed in Appendix~\ref{app:mezo}) executes these passes sequentially. However, these forward passes are inherently independent—the only difference being the random perturbations applied to the trainable parameters. To improve runtime under resource constraints while preserving the accuracy benefits of multi-query RGE, we propose the MobiZO framework. MobiZO enables parallel execution of gradient estimation through a specialized PEFT design called Multi-Perturbed LoRA (MP-LoRA). In addition, we show that MobiZO can be seamlessly integrated into inference engines such as ExecuTorch to enable practical and efficient on-device fine-tuning without modifying the runtime.

\textbf{MP-LoRA.}
A straightforward approach to performing multiple forward passes in parallel is to duplicate the model inputs and model weights, perturb each weight copy with distinct perturbations, and then execute the forward passes concurrently. However, this naive duplication incurs substantial memory overhead from replicating weights and managing perturbations, in terms of both storage and I/O access. The random seed trick introduced by MeZO~\cite{malladi2023finetuning} mitigates the memory overhead from storing the full perturbation vectors from $O(d)$ to $O(1)$. However, the computation time for the parameter perturbation step becomes $O(d)$ as trainable parameters must be sequentially updated using perturbations regenerated from the seed. This sequential process can substantially slow down training for large models, potentially negating the speedups gained from eliminating backpropagation.

To address both the memory overhead of parameter duplication and the $O(d)$ sequential parameter operations inherent in the seed trick, MP-LoRA leverages PEFT methods, which drastically reduce the number of trainable parameters. Our preliminary experiments (see Appendix~\ref{app:preliminary}) indicate that combining ZO with LoRA-FA yields superior performance compared to alternatives like DoRA~\cite{liu2024dora} and VeRA~\cite{kopiczko2024vera}. Consequently, LoRA-FA serves as the foundational PEFT method for our MP-LoRA design. We note that, while MP-LoRA is developed upon a LoRA-based PEFT method, its core principles can be adapted to other PEFT techniques.

\input{figures/1_method}

The MP-LoRA framework is designed to efficiently manage multiple perturbed states of LoRA parameters for concurrent forward passes. Similar to standard LoRA, MP-LoRA modules are applied to specific weights of the pre-trained model, augmenting them with a small number of trainable parameters. The MP-LoRA augmented model takes an input batch $\mathbf{x}$ and a set of $n$ distinct perturbations (which can be the full perturbation vectors or their generating seeds). At the beginning of a model execution, $n$ copies of the input batch are created, $\mathbf{X}^{(n)} = [\mathbf{x}_1, \mathbf{x}_2, \dots, \mathbf{x}_n]$. Each copy $\mathbf{x}_i$ is then associated with the $i$-th perturbation path for its entire traversal through the model's layers. This collection, $\mathbf{X}^{(n)}$, serves as the effective input that propagates through the network, encountering MP-LoRA layers. Within any given MP-LoRA layer, the original pre-trained model weights $\mathbf{W}$ and the LoRA $\mathbf{A}$ matrix are kept fixed and are shared across all $n$ operational paths passing through that layer. Only the LoRA $\mathbf{B}$ matrix is effectively replicated and distinctly perturbed $n$ times for these paths, resulting in a set of path-specific matrices $\mathbf{B}^{(n)} = [\mathbf{B}_1, \mathbf{B}_2, \dots, \mathbf{B}_n]$, where each $\mathbf{B}_i$ corresponds to the perturbation path $i$.

The output of an MP-LoRA layer for these $n$ concurrent paths is computed as:
$$\mathbf{Y}^{(n)} = \mathbf{X}^{(n)}\mathbf{W} + (\mathbf{X}^{(n)}\mathbf{A}) \odot \mathbf{B}^{(n)}$$
Here, $\odot$ denotes a batched operation where each component $(\mathbf{x}_i\mathbf{A})$ from the $n$ paths undergoes matrix multiplication with its respective perturbed LoRA matrix $\mathbf{B}_i$. Since $\mathbf{B}_i$ is orders of magnitude smaller than $\mathbf{W}$, the overall memory overhead is negligible even with the replication demands.  
MP-LoRA achieves its efficiency by ensuring these $n$ distinct path-specific computations per layer are performed concurrently, while reusing the shared $\mathbf{W}$ and $\mathbf{A}$ components.



\input{figures/3_executorch}

\textbf{Outer-Loop Parallelization.}
Each RGE step consists of evaluating $q$ distinct stochastic perturbations (queries). We employ MP-LoRA to execute these $q$ queries in parallel, as illustrated in Figure~\ref{fig:method}(a). This is achieved by invoking the MP-LoRA mechanism once with $n=q$ independent perturbations (or their random seeds) as input. This generates $q$ distinct perturbed model states that are subsequently processed concurrently through the model. However, performing $q$ operations simultaneously would naively increase the computational load per step by a factor of $q$.

To maintain a computational cost per training step comparable to that of single-query RGE, when performing MobiZO with $q > 1$ queries, we proportionally reduce the input batch size for each of the $q$ paths to $E = B/q$. Here, $B$ is the original batch size used in a $q=1$ setting, and $E$ is termed the \emph{effective batch size} per query path. For instance, if a baseline setting uses $q = 1$ and $B = 16$, our MobiZO approach can use $q = 4$ with an effective batch size $E = 4$ per query, thereby maintaining the total number of samples processed per step ($qE = B$). As we demonstrate in Section~\ref{sec:exp}, this trade-off of increasing $q$ while reducing $E$ often leads to improved model accuracy under a fixed computational budget. The motivation and theoretical background behind this trade-off are detailed in Appendix~\ref{app:trade-off}. Reducing the effective batch size $E$ in a multi-query setting also offers the ancillary benefit of reducing padding tokens, as shown in Appendix \ref{app:padding}. Typically, larger batch sizes lead to more padding, as sequences of varying lengths are padded to match the maximum sequence length within that larger batch. By adopting a smaller effective batch size per query, the total amount of padding can decrease, limiting wasted computation on these padding tokens, especially during attention operations.

Another key benefit of this MP-LoRA-enabled outer-loop parallelization is the improved data locality for model weights. By loading the shared weights (such as the base model weights $\mathbf{W}$ and LoRA $\mathbf{A}$ matrices within MP-LoRA layers) once and reusing them across all $q$ concurrent queries, costly external memory accesses are amortized. This can lead to a runtime per training step that is comparable to, or even faster than, the original sequential setting with $q=1$, despite processing multiple queries.

\textbf{Inner-Loop Parallelization.}
While outer-loop parallelization addresses concurrency across multiple queries, each gradient estimation in RGE still requires two forward passes per query: one with a positive perturbation and one with a negative perturbation. In standard RGE, these are typically executed sequentially.

To further accelerate each gradient estimation, MobiZO incorporates inner-loop parallelization, as outlined in Algorithm~\ref{alg:mobizo} (line 7). This is also enabled by MP-LoRA, which for a given query $i$, can perform both the positive and negative perturbation forward passes simultaneously, as illustrated in Figure~\ref{fig:method}(b). This is achieved by invoking MP-LoRA with $n=2q$, using perturbations $+\epsilon \mathbf{z}_i$ and $-\epsilon \mathbf{z}_i$ applied to the LoRA $\mathbf{B}$ matrix for the $i$-th path.
By processing positively and negatively perturbed states in a single MP-LoRA invocation, we obtain two corresponding outputs. The loss difference between these outputs can then be used to estimate the gradient component for that query, effectively performing the gradient estimation using a single forward pass. This approach further reduces the external memory bandwidth burden by maximizing the reuse of shared model weights across these two evaluations. Consequently, MobiZO with inner-loop parallelization can achieve an even faster runtime per training step compared to the sequential execution of two forward passes in RGE, even for $q=1$.

With inner-loop parallelization, the activation size at each layer is doubled, as it forwards two batches at the same time. However, this does not result in significant memory overhead. Unlike first-order methods, ZO methods allow activations from previous layers to be discarded during forward passes, rather than accumulating across all layers. This property, as noted in \cite{zhang2024revisiting}, enables ZO methods to scale more efficiently with long sequence lengths and large batch sizes compared to FO methods. To minimize memory costs for storing LoRA-$\mathbf{B}$ weight matrices, it is possible to keep a master copy of LoRA-$\mathbf{B}$ and instantiate perturbed copies dynamically during the forward pass. At each LoRA layer, only the master copy is updated with the gradient and learning rate. Perturbed copies of LoRA-$\mathbf{B}$ are then instantiated and deleted once the output is computed, ensuring that the number of additional trainable parameters remains the same as in the conventional ZO method.

A crucial benefit of MP-LoRA, enabling inner-loop parallelization in this manner, is that it forms the basis for on-device fine-tuning using inference engines, as detailed next.

\textbf{On-Device Training Adaptation.}
Deploying a model with inference engine ExecuTorch involves two primary steps: (1) converting a standard PyTorch \texttt{nn.Module} into an ExecuTorch program, a serialized computation graph with embedded parameters; and (2) offloading this binary file along with the C++ runtime to the edge device. The edge runtime then interprets and executes the model using a backend-specific operator library.

However, ExecuTorch does not natively support MeZO due to its reliance on complex device-side operations, such as random number generation, weight perturbation, and gradient application, none of which are exposed through standard ExecuTorch APIs. For instance, line 8 in Algorithm~\ref{alg:mobizo} requires direct modification of model weights, an operation currently unsupported. To address this limitation, we design a mobile version of the MP-LoRA that encapsulates all MobiZO logic inside its \texttt{forward} function. This approach enables full exportability and execution within the standard ExecuTorch runtime, eliminating the need to modify low-level components.

 Algorithm~\ref{alg:dual-forwarding} shows the \textbf{Mobile MP-LoRA} module for $q=1$. It maintains two perturbed variants of the LoRA weight matrix $\mathbf{B}$, each scaled by $\epsilon$ with positive and negative noise. Since ExecuTorch does not support resetting random seeds between forward passes, we store both perturbed versions of $\mathbf{B}$ in memory instead of regenerating them. During execution, the module computes the gradient using the difference between $\mathbf{B}[0]$ and $\mathbf{B}[1]$, restores the unperturbed weight, and updates the parameter via projected gradient. The final output is computed using the sum of the frozen linear term $\mathbf{x} \mathbf{W}$ and the adaptive term $\mathbf{x} \mathbf{A} \mathbf{B}$.

Figure~\ref{fig:executorch} illustrates the complete workflow. We begin with a pre-trained PyTorch model and replace its linear layers with our Mobile MP-LoRA modules. The modified model is then exported using the standard ExecuTorch pipeline and deployed to the edge device. On-device execution is handled entirely by the ExecuTorch runtime, which runs the fine-tuning process implicitly through the forward pass. Additionally, a lightweight noise generation operator is registered using the ExecuTorch extension API \cite{executorch-op} to support model perturbation.

\begin{algorithm}[h]
\caption{Mobile MP-LoRA Module}
\label{alg:dual-forwarding}
\begin{algorithmic}[1]
\STATE {\bfseries Input:} $\mathbf{x} \in \mathbb{R}^{2 \times \text{seq\_len} \times \text{k}}$, $\mathbf{A} \in \mathbb{R}^{k \times r}$, $\mathbf{B} \in \mathbb{R}^{2 \times r \times  k}$, $\mathbf{W}^{k \times  k}$, learning rate $\eta$, perturbation scale $\epsilon$, projected gradient $g$
\STATE $\text{diff} = \frac{\mathbf{B}[0] - \mathbf{B}[1]}{2}$
\STATE $\text{update} = \eta \cdot g \cdot \frac{\text{diff}}{\epsilon}$
\STATE $\mathbf{z} = \epsilon \cdot {randn\_like}(\mathbf{B}[0])$
\STATE $\mathbf{B}[0] = \mathbf{B}[0] - \text{diff} - \text{update} + \mathbf{z}$
\STATE $\mathbf{B}[1] = \mathbf{B}[1] + \text{diff} - \text{update} - \mathbf{z}$
\STATE $\text{output} = \mathbf{x}\mathbf{W} + bmm(\mathbf{x} \mathbf{A}, \mathbf{B})$ 
\STATE {\bfseries Return:} \text{output}
\end{algorithmic}
\end{algorithm}

\section{Experiments} 
\label{sec:exp}

\begin{table*}[]
\centering
\resizebox{0.8\textwidth}{!}{%
\begin{tabular}{llcccccc}
\cline{1-8}
\textbf{TinyLlama-1.1B} & \textbf{Methods $\backslash$ Tasks}   & \textbf{SST-2}& \textbf{RTE}   & \textbf{MRPC} & \textbf{QQP}  & \textbf{QNLI}  & \textbf{WNLI}    \\ \cline{1-8}
Zero-shot               & \multicolumn{1}{c}{} & 55.5           & 51.6          & 68.4           & 32.8          & 52.7          & 43.7             \\ \cline{1-8}
\multirow{2}{*}{FO-SGD} & Full                 & 93.0           & 80.6          & 80.0           & 84.0          & 83.6          & 58.2             \\
                        & LoRA-FA              & 93.0           & 77.7          & 79.1           & 83.3         & 84.3          & 54.9             \\ \cline{1-8}
\multirow{4}{*}{ZO-SGD} & MeZO (Full)          & \textbf{91.1}  & 65.3          & 70.8          & 73.7          & 69.2          & 58.2             \\
                        & MeZO (LoRA-FA)       & 86.5           & 67.1          & 72.1          & 74.7          & 62.4          & 59.2             \\
                        & MobiZO ($q=4$)        & 88.5           & 70.5          & \textbf{73.6} & 76.4          & 75.0          & \textbf{60.6}    \\
                        & MobiZO ($q=16$)       & 89.7           & \textbf{72.4} & 73.4          & \textbf{77.0} & \textbf{76.7} & 59.6             \\ \hline
\end{tabular}%
}
\end{table*}

\begin{table*}[]
\resizebox{\textwidth}{!}{%
\begin{tabular}{llcccccccccc}
\hline
\textbf{Llama2-7B}      & \textbf{Methods $\backslash$ Tasks}     & \textbf{SST-2} & \textbf{RTE}  & \textbf{BoolQ} & \textbf{WSC}  & \textbf{WiC}  & \textbf{MultiRC} & \textbf{COPA} & \textbf{WinoGrande} & \textbf{ARC-E} & \textbf{ARC-C}  \\ \hline
Zero-shot               &                      & 58.0           & 59.2          & 71.9           & 52.9          & 50.0          & 54.9             & 79.0          & 62.7          & 47.9           & 35.0       \\ \hline
\multirow{2}{*}{FO-SGD} & Full                 & 95.8           & 86.5          & 85.3           & 66.0          & 72.5          & 82.4             & 87.0          & 67.0          & 80.6           & 66.2       \\
                        & LoRA-FA              & 95.4           & 81.7          & 84.9           & 62.8          & 62.6          & 74.7            & 84.0          & 64.2          & 81.3           & 60.7       \\ \hline
ZO-SGD                  & MeZO (Full)          & 92.2           & 73.5          & 81.9           & 64.7          & 55.6          & 68.6            & 84.0          & 64.5          & 69.8          & 47.5        \\
                        & MeZO (LoRA-FA)       & 92.4           & 70.8          & 81.6           & 63.5          & 63.4          & 72.3            & 86.3          & 64.2          & 73.6          & 50.6        \\
                        & MobiZO ($q=4$)       & 94.1           & \textbf{75.5} & \textbf{82.7}  & \textbf{64.1} & 63.1          & \textbf{74.7}   & 85.0          & 64.8          & \textbf{73.8} & 50.9       \\
                        & MobiZO ($q=16$)       & \textbf{94.2}  & \textbf{75.5} & \textbf{82.7}  & 62.5         & \textbf{63.7} & 72.9            & \textbf{87.3} & \textbf{64.9} & 73.2          & \textbf{51.4}  \\ \hline
\end{tabular}%
}
\caption{Performance of fine-tuning TinyLlama-1.1B and Llama2-7B on different tasks with different optimizers. MobiZO outperforms the baseline MeZO in most tasks under the same computational budget.}
\label{tab:model-perf}
\end{table*}

We conduct comprehensive experiments on the TinyLlama-1.1B \cite{zhang2024tinyllama} and Llama2-7B \cite{touvron2023llama2} models across different systems to evaluate both fine-tuning performance and system efficiency. 

\subsection{Model Fine-Tuning Performance}

We compare two sets of baselines: the first employs an FO-SGD optimizer in both the full and LoRA-FA parameter spaces, while the second uses a ZO-SGD optimizer with MeZO ($q=1, B=16$) in the same parameter spaces. For our method, MobiZO, we ensure equivalent computation per training step while varying $q$ by scaling the effective batch size ($E$) to maintain a fixed $E*q$ value, such that setting $q=4, E=4$ or $q=16, E=1$ has equal computation load. By using the same number of training steps (i.e., 20,000) for both MobiZO and MeZO, we ensure that MobiZO does not exceed the computational budget of the MeZO baseline for end-to-end training.
Experiments are conducted using three random seeds, and we report the average performance. 
For reference, we also report zero-shot performance without additional fine-tuning. 

For the smaller-scale TinyLlama-1.1B model, we evaluate its performance on the GLUE dataset \cite{wang2019glue}. The results in Table \ref{tab:model-perf} show that increasing the number of queries while decreasing the batch size outperforms the baseline MeZO by up to $7.5\%$ accuracy. 
For the larger Llama2-7B model, we evaluate its performance on SST-2 \cite{wang2019glue}, SuperGLUE \cite{wang2020superglue}, WinoGrande \cite{sakaguchi2021winogrande}, ARC-Easy, and ARC-Challenge \cite{clark2018arc} datasets using the same experimental setup. Additional experimental details, including dataset descriptions, training procedures, and hyperparameters, are provided in Appendix \ref{app:exp}. 

From the results in Table~\ref{tab:model-perf}, we observe that MobiZO consistently outperforms MeZO across nearly all tasks on Llama2-7B model. Notably, MobiZO improves performance over the baseline that updates the full parameter space by up to $8.1\%$ in the WiC task, demonstrating its effectiveness in leveraging multi-query gradient estimation for improved fine-tuning quality under the same computational budget.
Although MobiZO introduces an additional hyperparameter $q$, we find that setting $q$ to 4 or 16 generally yields strong performance, reducing the need for extensive tuning. Additional results exploring a broader range of $q$ values are provided in Appendix~\ref{app:addtioanl_zo}, along with an in-depth discussion of the accuracy gains enabled by MobiZO’s design choices.

\subsection{System Performance}
\label{sec:server-side}
We conduct measurements on a single NVIDIA A100 GPU to evaluate the server-side system performance of MobiZO compared to its baselines. The ZO-SGD optimizer, including both MeZO and MobiZO, performs forward passes in 16-bit floating-point precision to maximize computational efficiency, leveraging ZO's tolerance for low-precision gradient estimation \cite{zhang2024revisiting}. We use the FO-SGD optimizer with mixed-precision training enabled for memory and runtime evaluations.

\textbf{Memory Efficiency.} We first evaluate the peak memory usage of MobiZO across different fixed sequence length and batch size configurations. The reported memory footprint includes storage for weights, activations, gradients, CUDA kernels, and other implementation-specific details.

\begin{table}[]
\resizebox{\columnwidth}{!}{%
\begin{tabular}{lccccccccc}
\hline
\multicolumn{1}{c}{Sequence length} &       & 64    &       &       & 128   &       &       & 256   &       \\ \hline
\multicolumn{1}{c}{Batch size}      & 1     & 8     & 16    & 1     & 8     & 16    & 1     & 8     & 16    \\ \hline
\multicolumn{10}{c}{\textbf{\textit{TinyLlama-1.1B}}}                                                                         \\ \hline
FO (Full)                           & 11.32 & 12.00 & 12.78 & 11.44 & 13.00 & 14.76 & 11.77 & 15.62 & 20.02 \\
FO (LoRA-FA)                        & 4.15  & 5.00  & 5.98  & 4.27  & 5.98  & 7.81  & 4.51  & 7.81  & 11.58 \\
MeZO (LoRA-FA)                      & 2.09  & 2.19  & 2.32  & 2.10  & 2.32  & 2.56  & 2.13  & 2.56  & 3.05  \\
MobiZO                               & 2.11  & 2.32  & 2.56  & 2.14  & 2.56  & 3.05  & 2.20  & 3.05  & 3.98  \\ \hline
\multicolumn{10}{c}{\textbf{\textit{Llama2-7B}}}                                                                              \\ \hline
FO (Full)                           & 64.31 & 66.12 & 69.20 & 64.6  & 68.51 & 72.97 & 65.32 & 74.22 & 84.40 \\
FO (LoRA-FA)                        & 25.16 & 27.20 & 29.58 & 25.46 & 29.58 & 34.28 & 26.05 & 34.29 & 43.66 \\
MeZO (LoRA-FA)                      & 12.59 & 12.70 & 12.82 & 12.61 & 12.82 & 13.06 & 12.64 & 13.06 & 13.55 \\
MobiZO                               & 12.61 & 12.82 & 13.06 & 12.64 & 13.07 & 13.55 & 12.70 & 13.55 & 14.53 \\ \hline
\end{tabular}%
}
\caption{Peak memory usage (GB) of TinyLlama-1.1B and Llama2-7B for different sequence length and batch size configurations.}
\label{tab:memory}
\end{table}

Table \ref{tab:memory} shows the memory usage of FO-SGD (LoRA-FA), MeZO (LoRA-FA), and MobiZO. The FO-SGD optimizer requires more memory due to storing activations from all intermediate layers, despite minimal gradient and optimizer state storage through PEFT. In contrast, MobiZO slightly increases memory usage due to the increased size of the largest output tensor during the forward pass and instantiation of multiple sets of LoRA trainable parameters, yet it still demands significantly less memory than the FO optimizer. For instance, with Llama2-7B, a sequence length of 256, and a batch size of 16, memory usage increases from 13.55 GB to 14.53 GB for MobiZO, whereas FO requires over 40 GB. FO over the full parameter space requires even much more memory, going beyond the memory capacity of edge devices.

\input{figures/4_end2end_time}

\textbf{End-to-end Wall-clock Time Speedup.} Figure \ref{fig:end2end-time} shows the end-to-end wall-clock time for fine-tuning TinyLlama-1.1B and Llama2-7B using MeZO and MobiZO for 20,000 steps across various tasks. By applying PEFT methods, both MeZO and MobiZO reduce training time by minimizing sequential processing of model parameters, a benefit that becomes more pronounced with larger models such as Llama2-7B. MobiZO further improves training runtime through inner-loop and outer-loop parallelization achieving speedups of up to $4.3\times$ over MeZO (Full) and up to $1.9\times$ over MeZO (LoRA-FA) baselines.

Additional system profiling ablation studies, including runtime breakdown under different fixed sequence length and batch size configurations, as well as under different quantization schemes, are available in Appendix \ref{app:ablation}.

\subsection{On-Device Training Experiments}

For on-device training experiments, we begin with a sanity check to verify per-step loss values on two edge platforms: the NVIDIA Jetson Nano Orin (8GB) GPU and the OnePlus 12 smartphone (12GB) NPU backend. This ensures that both platforms yield the same output given the same input as those observed on the server side. Detailed edge system specifications and experimental setups are provided in Appendix \ref{app:edge}. After verification, we measure and report the runtime per step of MobiZO across different fixed sequence lengths and batch size configurations, following the same setup in Section \ref{sec:server-side}. Due to out-of-memory issues, FO training is omitted from on-device experiments.

\begin{table}[]
\resizebox{\columnwidth}{!}{%
\begin{tabular}{lcccccccc}
\hline
\multicolumn{1}{c}{Sequence length} &       & 64    &       &       & 128   &       &               \\ \hline
\multicolumn{1}{c}{Batch size}      & 1     & 2     & 4     & 8     & 1     & 2     & 4     & 8     \\ \hline
\multicolumn{9}{c}{\textbf{\textit{TinyLlama-1.1B}}}                                                \\ \hline
MeZO (LoRA-FA)                      & 0.69  & 0.71  & 0.89  & 1.28  & 0.70  & 0.88  & 1.27  & 2.18  \\
MobiZO                               & 0.43  & 0.49  & 0.69  & 1.15  & 0.49  & 0.69  & 1.13  & 2.00  \\
Speedup ratio                       & 1.62  & 1.45  & 1.29  & 1.12  & 1.42  & 1.29  & 1.12  & 1.09  \\ \hline
\multicolumn{9}{c}{\textbf{\textit{Llama2-7B}}}                                                     \\ \hline
MeZO (LoRA-FA)                      & 3.10  & 3.37  & 4.44  & 6.46  & 3.37  & 4.44  & 6.47  & 10.83 \\
MobiZO                               & 1.69  & 2.22  & 3.22  & 5.38  & 2.22  & 3.22  & 5.37  & 8.60  \\
Speedup ratio                       & 1.83  & 1.52  & 1.38  & 1.20  & 1.52  & 1.38  & 1.21  & 1.26  \\ \hline
\end{tabular}%
}
\caption{Runtime (sec/step) and speedup ratio of inner-loop parallelization on Jetson GPU backend for TinyLlama-1.1B and Llama2-7B with NF4 quantization.}
\label{tab:jetson}
\end{table}

On the Jetson platform, which runs on a Linux system, we use the PyTorch library for model forward passes. Table \ref{tab:jetson} demonstrates the speedup achieved by MobiZO with inner-loop parallelization and NF4 weight-only quantization, reaching up to 1.83$\times$ improvement by eliminating the repeated weight dequantization overhead. Since MobiZO is fully compatible with Q-LoRA~\cite{dettmers2023qlora}, we also verify that weight-only quantization does not significantly degrade accuracy, as demonstrated in Appendix~\ref{app:addtioanl_zo}.

On the smartphone platform, which operates on Android OS without PyTorch support, we repurpose the latest ExecuTorch library (v0.6.0) to perform MobiZO fine-tuning through integrating the Mobile MP-LoRA module as described in Section \ref{sec:method}. Since we do not modify the runtime code on the edge device, vanilla MeZO baseline experiments are omitted due to incompatibility. Additionally, due to current limitations in ExecuTorch’s support for weight-only quantization, we run TinyLlama-1.1B in FP16 mode on the NPU backend. While Qualcomm's NPU is optimized for low power rather than raw throughput \cite{qualcomm_hexagon_sdk}, Executorch on the Android platform achieves comparable efficiency to Pytorch on the Jetson CUDA platform at smaller batch sizes. As shown in Figure \ref{fig:NPU}, with an effective batch size of 16, the NPU backend takes 5.76 seconds for one step with a sequence length of 128, whereas Jetson GPU completes it in 2.00 seconds.

\input{figures/5_NPU}

\section{Conclusion}
This work presents MobiZO, a parallelized zeroth-order optimization framework for efficient fine-tuning of large language models in resource-constrained environments. By combining outer-loop and inner-loop parallelization with the Multi-Perturbed LoRA design, MobiZO removes the sequential bottleneck of multi-query gradient estimation while retaining its accuracy benefits. Experiments on server GPUs, Jetson platforms, and Android NPUs show that MobiZO reduces wall-clock training time, lowers memory usage, and improves fine-tuning accuracy over baselines such as MeZO. Its seamless integration with inference engines like ExecuTorch requires no runtime modifications, enabling practical real-time on-device fine-tuning. Overall, MobiZO bridges the gap between zeroth-order optimization and real-world deployment, paving the way for trustworthy and personalized LLMs on mobile and edge devices.

\section{Limitations}
While MobiZO enables efficient on-device LLM fine-tuning, it has several limitations.
First, MobiZO is tailored for the randomized gradient estimator in ZO optimization. Extending it to other ZO methods, such as variance-reduced optimizers or adaptive query selection, could further improve convergence speed.
Second, thermal constraints on edge devices limit sustained compute performance, leading to throttling during prolonged fine-tuning. Future work will explore thermal-aware scheduling to maintain performance under temperature fluctuations.
Third, MobiZO assumes static computational settings, whereas edge environments often have dynamic compute and memory resource constraints. Adapting query count, batch size, or precision in response to runtime conditions is a promising direction.

\section*{Acknowledgment}
We sincerely thank all the reviewers for their time and constructive comments. This material is based upon work supported by  NSF award numbers 2224319 and 1956435, REAL@USC-Meta center, and VMware gift. The views, opinions, and/or findings expressed are those of the author(s) and should not be interpreted as representing the official views or policies of the  U.S. Government.  We would also like to extend our thanks to the ExecuTorch team from Meta and Yanzhi Wang from Northeastern University for
their valuable discussions and advice.

\bibliography{custom}

\clearpage
\appendix

\section{MeZO Algorithm and Its Limitation}
\label{app:mezo}

\begin{algorithm}[h]
\caption{MeZO with $q>1$.}
\label{alg:mezo}
\begin{algorithmic}[1]
\STATE {\bfseries Input:} parameters $\bm{\theta} \in \mathbb{R}^d$, loss $\mathcal{L}: \mathbb{R}^d \rightarrow \mathbb{R}$, step budget $T$, function query budget $q$, perturbation scale $\epsilon$, learning rate $\eta$
\FOR{$t = 1, \dots, T$}
    \FOR{$i = 1, \dots, q$}
        \STATE \texttt{seeds, projected\_grads} $=$ \texttt{[]}
        \STATE Sample batch $\bm{\mathcal{B}} \subset \mathcal{D}$ and random seed $s$
        \STATE $\bm{\theta} =$ \texttt{PerturbParameters}($\bm{\theta}, \epsilon, s$)
        \STATE $\ell_+ = \mathcal{L}(\bm{\theta}; \bm{\mathcal{B}})$
        \STATE $\bm{\theta} =$ \texttt{PerturbParameters}($\bm{\theta}, -2\epsilon, s$)
        \STATE $\ell_- = \mathcal{L}(\bm{\theta}; \bm{\mathcal{B}})$
        \STATE $\bm{\theta} =$ \texttt{PerturbParameters}($\bm{\theta}, \epsilon, s$)
        \STATE \texttt{proj\_grads[i]} $= \frac{\ell_+ - \ell_-}{2\epsilon}$
        \STATE \texttt{seeds[i]} $= s$
    \ENDFOR
    \FOR{$i = 1, \dots, q$}
        \STATE Reset random generator with \texttt{seeds[i]}
        \FOR{$\bm{\theta}_j \in \bm{\theta}$}
            \STATE $z \sim \mathcal{N}(0, 1)$
            \STATE $\bm{\theta}_j = \bm{\theta}_j - \frac{\eta_t}{q} \times \texttt{proj\_grads[i]} \times z$
        \ENDFOR
    \ENDFOR
\ENDFOR
\STATE \textbf{Function} {\texttt{PerturbParameters}($\bm{\theta}, \epsilon, s$)}
    \STATE Reset random number generator with seed $s$
    \FOR{$\bm{\theta}_j \in \bm{\theta}$}
        \STATE $z \sim \mathcal{N}(0, 1)$
        \STATE $\bm{\theta}_j = \bm{\theta}_j + \epsilon z$
    \ENDFOR
\STATE \textbf{End Function}
\end{algorithmic}
\end{algorithm}

We evaluate the runtime efficiency of the MeZO optimizer, outlined in Algorithm~\ref{alg:mezo}, which is adapted from the original work. MeZO employs a random seed trick to eliminate the need for storing random noise, reducing peak memory usage.

In each iteration, MeZO proceeds through four distinct loops. First, it introduces positive noise into the trainable parameters (line 6), followed by perturbing the weights in the opposite direction using the same noise (line 8). Next, the weights are restored to their original state before the update (line 10), and finally, the computed gradients are applied to update the weights (line 18).

This method reduces memory overhead from $O(d)$ to $O(1)$ by avoiding the storage of random noise. However, the runtime cost escalates from $O(1)$ to $O(d)$ because each parameter update requires individual processing, which cannot be efficiently parallelized. In practical settings, especially with LLMs, iterating over the full parameter set four times per update can significantly slow down the training process, thus negating the benefits of eliminating backpropagation.

In contrast, PyTorch's FO optimizers utilize a \textit{foreach} implementation by default. This method aggregates all layer weights into a single tensor during parameter updates, which speeds up the computation. However, this approach also increases the memory usage by $O(d)$, as it requires maintaining a copy of the entire gradients for the parameter update.

Table~\ref{tab:mezo_runtime} compares the runtime of the Llama2-7B model using both FO-SGD and MeZO-SGD optimizers ($q=1$) over the full parameter space across various batch sizes and sequence lengths on the same standard benchmark introduced in Section \ref{sec:server-side}. The FO optimizer is run with FP16 mixed-precision training, while MeZO uses pure FP16 to maximize computational speed. To avoid out-of-memory errors, we utilize two NVIDIA A100 (40GB) GPUs for the FO optimizer, which incurs additional GPU communication time in a distributed environment.

\begin{table}[h]
\centering
\resizebox{\columnwidth}{!}{
\begin{tabular}{llllllllll}
\hline
Sequence length       & \multicolumn{3}{c}{64} & \multicolumn{3}{c}{128} & \multicolumn{3}{c}{256} \\ \hline
Batch size            & 1      & 4     & 8     & 1      & 4      & 8     & 1      & 4      & 8     \\ \hline
FO-SGD                & 0.17   & 0.21  & 0.34  & 0.19   & 0.33   & 0.49  & 0.18   & 0.49   & 0.90  \\ \hline
MeZO-SGD ($q=1$)      & 0.43   & 0.48  & 0.56  & 0.43   & 0.56   & 0.73  & 0.45   & 0.73   & 1.05  \\ \hline
\end{tabular}
}
\caption{Runtime (sec/step) of Llama2-7B using FO and MeZO optimizers over full parameter space.}
\label{tab:mezo_runtime}
\end{table}

When both the batch size and sequence length are small, MeZO exhibits significantly higher runtime due to the overhead of sequential operations required to apply perturbations and gradients. However, as the batch size and sequence length increase, where forward and backward passes, as well as GPU communication, dominate the runtime, the MeZO optimizer demonstrates improved performance. This behavior highlights the importance of applying PEFT methods with MeZO to mitigate the computation overhead caused by the sequential processing of model parameters.

\section{Preliminary Experiment of ZO with Different PEFT Methods}
\label{app:preliminary}

We conducted a preliminary experiment by fine-tuning the OPT-1.3B model \cite{zhang2022opt} for 10,000 iterations on the SST2 dataset \cite{wang2019glue} using ZO-SGD optimizer with different PEFT methods. We use hyperparameter grid search with learning rate $\in \{5e-6, 5e-5, 5e-4, 5e-3 \}$ and $\epsilon \in \{1e-3, 1e-2\}$. LoRA \cite{hu2022lora}, LoRA-FA \cite{zhang2023lorafa}, and DoRA \cite{liu2024dora} are configured with $r=16$, and VeRA \cite{kopiczko2024vera} uses $r=1024$.
The results in Table \ref{tab:preliminary} indicate that the LoRA-FA method outperforms other PEFT methods in terms of accuracy.

\begin{table}[h]
\centering
\resizebox{\columnwidth}{!}{%
\begin{tabular}{lllll}
\hline
\textbf{PEFT Methods}    & LoRA  & LoRA-FA & DoRA & VeRA  \\ \hline
\textbf{Accuracy}        & 90.9  & 92.0    & 90.9 &   91.4    \\ \hline
\end{tabular}
}
\caption{ZO accuracy of OPT-1.3B on SST2 dataset using different PEFT methods.}
\label{tab:preliminary}
\end{table}

\section{Padding Statistics}
\label{app:padding}

Figure \ref{fig:padding} illustrates how smaller batch sizes (e.g., 2) result in less padding compared to larger batch sizes (e.g., 8), thereby reducing wasted computation.

\input{figures/2_padding}

Figure \ref{fig:padding_tokens} shows the average percentage of padding tokens used across different tasks and batch sizes. A larger batch size of 16 results in a higher percentage of padding tokens across all tasks compared to a batch size of 4. This suggests that smaller batch sizes may help reduce padding overhead, potentially leading to more efficient computation.

\input{figures/app_padding_stats}

\section{Experiment Setup}
\label{app:exp}

\subsection{Datasets}
We evaluate the performance of the TinyLlama-1.1B model on six tasks from the GLUE dataset \cite{wang2019glue}: sentiment analysis (SST2), paraphrase (MRPC and QQP), and natural language inference (QNLI, RTE, and WNLI).
For the larger Llama2-7B model, evaluations were performed on two tasks from the GLUE dataset: SST2 and RTE. Additionally, the model was tested on six tasks from the SuperGLUE dataset \cite{wang2020superglue}, categorized as follows: text classification (BoolQ, WSC, WIC, and MultiRC), and multiple-choice (COPA). We include three additional multiple-choice tasks from the WinoGrande \cite{sakaguchi2021winogrande}, ARC-Easy, and ARC-Challenge \cite{clark2018arc} datasets.
For question-and-answering tasks, we utilize the F1 score as a metric, while accuracy metrics are used for the rest. All datasets used in this work are in English.

\begin{table*}[t]
\centering
\resizebox{0.7\textwidth}{!}{%
\begin{tabular}{lll}
\hline
\textbf{Dataset} & \textbf{Type} & \multicolumn{1}{c}{\textbf{Prompt}}  \\ \hline
SST-2            & cls.          & \texttt{<text>} It was terrible/great    \\ \hline
RTE              & cls.          & \texttt{<premise>} Does this mean that ``\texttt{<hypothesis>}'' is true? Yes or No? \\
                 &               & Yes/No  \\ \hline
MRPC             & cls.          & Do the following two sentences mean the same thing? Yes or No? \\
                 &               & Sentence 1: \texttt{<sentence1>} \\
                 &               & Sentence 2: \texttt{<sentence2>} \\
                 &               & Yes/No  \\ \hline
QQP              & cls.          & Are these two questions asking the same thing? Yes or No? \\
                 &               & Question 1: \texttt{<question1>} \\
                 &               & Question 2: \texttt{<question2>} \\
                 &               & Yes/No  \\ \hline
QNLI             & cls.          & Does this sentence answer the question? Yes or No? \\
                 &               & Sentence 1: \texttt{<sentence1>} \\
                 &               & Sentence 2: \texttt{<sentence2>} \\
                 &               & Yes/No  \\ \hline
WNLI             & cls.          & Given the first sentence, is the second sentence true? Yes or No? \\
                 &               & Sentence 1: \texttt{<sentence1>} \\
                 &               & Sentence 2: \texttt{<sentence2>} \\
                 &               & Yes/No  \\ \hline
BoolQ            & cls.          & \texttt{<passage>} \texttt{<question>} \texttt{<answer>}? \\
                 &               & Yes/No  \\ \hline
WSC              & cls.          & \texttt{<text>} In the previous sentence, does the pronoun ``\texttt{<span2>}'' refer to \texttt{<span1>}? \\
                 &               & Yes/No  \\ \hline
WIC              & cls.          & Does the word ``\texttt{<word>}'' have the same meaning in these two sentences? \\
                 &               & \texttt{<sent1>} \texttt{<sent2>} \\
                 &               & Yes, No?  \\ \hline
MultiRC          & cls.          & \texttt{<paragraph>} Question: \texttt{<question>} \\
                 &               & I found this answer ``\texttt{<answer>}''. Is that correct? \\
                 &               & Yes or No?  \\ \hline
COPA             & mch.          & \texttt{<premise>} so/because \texttt{<candidate>}  \\ \hline
WinoGrande       & mch.          & \texttt{<context>} \texttt{<subject>} \texttt{<object>} \\ \hline
ARC              & cls.          & Pick the most correct option to answer the following question: \texttt{<question>} \\
                 &               & Options: \texttt{<op1>, <op2>, <op3>, <op4>} \\
                 &               & Answer: \texttt{<label>} \\
\hline
\end{tabular}%
}
\caption{The prompt template of the datasets used in the experiments.}
\label{tab:template}
\end{table*}

\subsection{Training procedure}
We achieve text classification, multiple-choice, and question-and-answering tasks through next-word prediction, using prompt templates based on MeZO \cite{malladi2023finetuning} and PromptSource \cite{bach2022promptsource}. 
Table \ref{tab:template} presents the prompt templates used for the datasets in our TinyLlama-1.1B and Llama2-7B experiments. For SST-2, RTE, BoolQ, WSC, WIC, MultiRC, and COPA, we applied the template from MeZO \cite{malladi2023finetuning}. We created templates for MRPC, QQP, QNLI, WNLI, and ARC by following the suggestions from PromptSource \cite{bach2022promptsource}, and we adapted the same template for WinoGrande from \cite{zhang2024revisiting}.

Unlike MeZO, we compute the loss value of prediction over the entire vocabulary space instead of only the vocabulary space of the ground true.
For these tests, we also adopt a low-volume data condition, limiting our samples to 1,000 for training, 500 for validation, and 1,000 for testing, as proposed in the original MeZO work \cite{malladi2023finetuning}.
FO-SGD experiments are trained for 1,000 iterations, and performance on the test dataset is evaluated every 100 steps. ZO experiments are trained for 20,000 iterations and performance on the test dataset is evaluated every 500 steps.

\subsection{Hyperparameters}

We report the hyperparameter search grids in Table~\ref{tab:hyperparameters}. For LoRA hyperparameters, we choose the LoRA rank to be 16 and the LoRA alpha to be 32. For MobiZO, with the constant batch size of 16, we search configurations ($q = 1, E = 16$), ($q = 2, E = 8$), ($q = 4, E = 4$), ($q = 8, E = 2$), and ($q = 16, E = 1$). Note that MeZO (LoRA-FA) is a special case of MobiZO with $q=1$.

\section{Additional FO Experiments}

We also provide additional experimental results on FO-Adam in Tables \ref{tab:fo1} and \ref{tab:fo2}. While FO-Adam can enhance model performance, it introduces a significantly higher memory overhead, particularly when updating all model parameters. This is because Adam maintains two state variables, moment estimates of the first and second order, for each parameter, effectively tripling the memory requirement compared to storing only the model parameters. Therefore, FO-Adam is typically deployed in distributed multi-GPU environments, which further increases runtime due to the overhead of inter-device communication. 

\begin{table}[h!]
\centering
\resizebox{\columnwidth}{!}{%
\begin{tabular}{llcccccc}
\hline
\textbf{Methods}   & \textbf{SST-2} & \textbf{RTE} & \textbf{MRPC} & \textbf{QQP} & \textbf{QNLI} & \textbf{WNLI} \\ \hline
Full      & 91.9           & 72.5         & 77.4          & 82.4         & 80.8          & 56.3          \\
LoRA-FA   & 94.2           & 82.6         & 82.3          & 84.4         & 86.5          & 56.3          \\ \hline
\end{tabular}
}
\caption{Performance of fine-tuning TinyLlama-1.1B on different tasks with FO-Adam optimizers.}
\label{tab:fo1}
\end{table}

\begin{table}[h!]
\centering
\resizebox{\columnwidth}{!}{%
\begin{tabular}{llcccccccccc}
\hline
\textbf{Methods}   & \textbf{SST-2} & \textbf{RTE} & \textbf{BoolQ} & \textbf{WSC} & \textbf{WiC} & \textbf{MultiRC} & \textbf{COPA} \\ \hline
Full             & 92.5           & 78.7         & 80.6           & 63.4         & 67.2         & 71.7             & 81.0                 \\
LoRA-FA          & 96.0           & 88.1         & 85.7           & 79.8         & 75.1         & 84.2             & 87.0                 \\ \hline
\end{tabular}
}
\caption{Performance of fine-tuning Llama2-7B on different tasks with FO-Adam optimizers.}
\label{tab:fo2}
\end{table}

\begin{table*}[t]
\centering
\resizebox{0.8\textwidth}{!}{%
\begin{tabular}{lllll}
\hline
\multicolumn{1}{l}{\textbf{Methods}} & \multicolumn{2}{c}{\textbf{TinyLlama-1.1B}} & \multicolumn{2}{c}{\textbf{Llama2-7B}}     \\ \hline
FO (Full)    & Batch size    & 16                   & Batch size    & 8                    \\
             & Learning rate & \{1e-4, 5e-5, 1e-5\} & Learning rate & \{1e-4, 5e-5, 1e-5\} \\ \hline
FO (LoRA-FA) & Batch size    & 16                   & Batch size    & 8                    \\
             & Learning rate & \{5e-3, 1e-3, 5e-4\} & Learning rate & \{5e-3, 1e-3, 5e-4\} \\ \hline
MeZO (Full)  & Batch size    & 16                   & Batch size    & 16                   \\
             & Learning rate & \{1e-6, 5e-7, 1e-7\} & Learning rate & \{1e-6, 5e-7, 1e-7\} \\
             & $\epsilon$    & 1e-3                 & $\epsilon$    & 1e-3                 \\ \hline
MobiZO       & Batch size    & 16                   & Batch size    & 16                   \\
             & $q$           & \{1, 2, 4, 8, 16\}   & $q$           & \{1, 2, 4, 8, 16\}   \\
                              & Learning rate  & \{5e-4, 1e-4, 5e-5, 1e-5\} & Learning rate & \{5e-4, 1e-4, 5e-5, 1e-5\} \\
             & $\epsilon$    & 1e-2                 & $\epsilon$    & 1e-2                 \\ \hline
\end{tabular}
}
\caption{Hyperparameters used for TinyLlama-1.1B and Llama2-7B experiments.}
\label{tab:hyperparameters}
\end{table*}

\section{Additional ZO Experiments}
\label{app:addtioanl_zo}

\subsection{Legacy OPT models} 
We conducted additional zeroth-order optimization experiments on the OPT-1.3B model \cite{zhang2022opt}, which was also evaluated in the original MeZO study~\cite{malladi2023finetuning}. Given our emphasis on deployment in resource-constrained edge environments, we exclude larger models such as OPT-13B and OPT-70B from consideration. As shown in Table~\ref{tab:opt}, MobiZO consistently outperforms MeZO across most tasks, highlighting its effectiveness beyond the LLaMA model family. These results complement our primary evaluations on LLaMA-based models and further demonstrate the general applicability of MobiZO across diverse model architectures.

\begin{table}[h]
\centering
\resizebox{\columnwidth}{!}{%
\begin{tabular}{lcccccc}
\hline
\textbf{Methods}        & \textbf{SST-2} & \textbf{RTE} & \textbf{MRPC} & \textbf{QQP} & \textbf{QNLI} & \textbf{WNLI} \\
\hline
MeZO (Full)             & \textbf{92.4}  & \underline{57.8} & \underline{70.6} & 67.7 & 56.2 & \underline{59.2} \\
MeZO (LoRA-FA)          & 89.8  & \textbf{62.8} & \underline{70.6} & 69.3 & 58.9 & \underline{59.2} \\
MobiZO ($q = 4$)         & \underline{92.0}  & \textbf{62.8} & \textbf{71.3} & \underline{73.1} & \underline{65.4} & \textbf{60.6} \\
MobiZO ($q = 16$)        & 91.5  & \textbf{62.8} & \textbf{71.3} & \textbf{74.3} & \textbf{65.8} & \textbf{60.6} \\
\hline
\end{tabular}
}
\caption{Evaluation results on GLUE tasks using OPT-1.3B with different ZO methods.}
\label{tab:opt}
\end{table}

\subsection{Latest edge-specific models} 

We further evaluate three latest sub-1B parameter models: Llama3.2-1B \cite{meta2024llama3}, Qwen3-0.6B \cite{yang2025qwen3technicalreport}, and SmolLM-340M \cite{yang2025qwen3technicalreport}. These models fit within 6–8 GB memory and are representative of mobile and edge platforms. Results in Tables \ref{tab:llama3}, \ref{tab:qwen3}, and \ref{tab:smollm} show that MobiZO’s advantages extend to other architectures and smaller LLMs. MobiZO consistently improves task performance while also benefiting from the training speedup of parallelized execution.

\begin{table}[h]
\centering
\resizebox{\columnwidth}{!}{%
\begin{tabular}{lcccccc}
\hline
\textbf{Methods}         & \textbf{SST-2} & \textbf{RTE} & \textbf{MRPC} & \textbf{QQP} & \textbf{QNLI} & \textbf{WNLI} \\
\hline
MeZO (Full)              & 91.9  & 63.9 & 70.6 & 75.6 & \underline{67.7} & \underline{60.6} \\
MeZO (LoRA-FA)           & 92.8  & \underline{64.6} & 70.3 & 73.6 & 60.4 & \textbf{63.4} \\
MobiZO ($q = 4$)          & \textbf{93.8}  & \underline{64.6} & \textbf{74.8} & \underline{76.2} & 63.7 & \textbf{63.4} \\
MobiZO ($q = 16$)         & \underline{93.7}  & \textbf{65.3} & \underline{73.0} & \textbf{77.9} & \textbf{67.8} & \textbf{63.4} \\
\hline
\end{tabular}
}
\caption{Evaluation results on GLUE tasks using Llama3.2-1B with different ZO methods.}
\label{tab:llama3}
\end{table}

\begin{table}[h]
\centering
\resizebox{\columnwidth}{!}{%
\begin{tabular}{lcccccc}
\hline
\textbf{Methods}          & \textbf{SST-2} & \textbf{RTE} & \textbf{MRPC} & \textbf{QQP} & \textbf{QNLI} & \textbf{WNLI} \\
\hline
MeZO (Full)               & \underline{87.6} & 74.7 & 74.3 & 76.9 & 66.3 & \textbf{69.0} \\
MeZO (LoRA-FA)            & 87.2 & \textbf{77.9} & 78.7 & 79.1 & 73.6 & 63.4 \\
MobiZO ($q = 4$)          & \textbf{87.8} & \underline{75.8} & \textbf{79.2} & \underline{80.5} & \underline{74.6} & \underline{66.2} \\
MobiZO ($q = 16$)         & 87.4 & 75.5 & \underline{78.9} & \textbf{81.1} & \textbf{80.8} & 63.4 \\
\hline
\end{tabular}
}
\caption{Evaluation results on GLUE tasks using Qwen3-0.6B with different ZO methods.}
\label{tab:qwen3}
\end{table}

\begin{table}[h!]
\centering
\resizebox{\columnwidth}{!}{%
\begin{tabular}{lcccccc}
\hline
\textbf{Methods}          & \textbf{SST-2} & \textbf{RTE} & \textbf{MRPC} & \textbf{QQP} & \textbf{QNLI} & \textbf{WNLI} \\
\hline
MeZO (Full)               & 75.0 & 55.2 & 68.6 & 67.3 & 55.4 & \textbf{57.8} \\
MeZO (LoRA-FA)            & \textbf{87.6} & 54.2 & \underline{72.8} & \underline{71.8} & 73.2 & 53.5 \\
MobiZO ($q = 4$)          & \underline{87.4} & \underline{56.0} & \textbf{73.5} & 71.1 & \textbf{75.3} & \underline{56.3} \\
MobiZO ($q = 16$)         & \underline{87.4} & \textbf{57.4} & \underline{72.8} & \textbf{74.2} & \underline{74.6} & 53.5 \\
\hline
\end{tabular}
}
\caption{Evaluation results on GLUE tasks using SmolLM-340M with different ZO methods.}
\label{tab:smollm}
\end{table}

\subsection{MobiZO with weight-only quantization}

To evaluate the compatibility of MobiZO with quantized models, we apply weight-only quantization at both 8-bit and 4-bit precision levels to the TinyLlama-1.1B model, building on Q-LoRA~\cite{dettmers2023qlora}, which has shown that NF4 quantization is effective for freezing pretrained weights. In this setup, weights are stored in low-bit formats to reduce memory usage and are dequantized to FP16 during the forward pass. As shown in Table~\ref{tab:qlora} and Table~\ref{tab:model-perf}, the fine-tuning performance remains comparable to that of full FP16 training, confirming that low-bit quantization does not significantly affect model accuracy.

\begin{table}[h]
\centering
\resizebox{\columnwidth}{!}{%
\begin{tabular}{lcccccc}
\hline
\textbf{Methods}         & \textbf{SST-2} & \textbf{RTE} & \textbf{MRPC} & \textbf{QQP} & \textbf{QNLI}  &  \textbf{WNLI} \\ \hline
\multicolumn{7}{l}{\textit{8-bit weights}} \\
MobiZO ($q=4$)           & 88.3       & 70.3     & 73.4     & 76.2     & 74.8     & 60.3 \\
MobiZO ($q=16$)          & 89.4       & 72.2     & 73.5     & 76.9     & 76.5     & 59.7 \\ \hline
\multicolumn{7}{l}{\textit{4-bit weights}} \\
MobiZO ($q=4$)           & 88.0       & 70.1     & 73.2     & 76.1     & 74.6     & 59.6 \\
MobiZO ($q=16$)          & 88.9       & 72.1     & 73.0     & 77.1     & 76.4     & 58.2 \\ \hline
\end{tabular}
}
\caption{Performance of TinyLlama-1.1B on GLUE tasks under 8-bit and 4-bit weight-only quantization.}
\label{tab:qlora}
\end{table}

\section{In-depth analysis of trade-offs in RGE} \label{app:trade-off}

Table \ref{tab:trade-off} summarizes different trade-offs in RGE under a fixed computational budget. One strategy (Row 2) suggested by MeZO compensates for the increased number of queries by reducing the total number of training steps.
In this work, we introduce an alternative trade-off (Row 3): increasing the query count while decreasing the batch size, rather than reducing training steps. As demonstrated in Section \ref{app:exp}, this approach consistently outperforms the trade-offs in Rows 1 and 2. Nonetheless, each training step takes longer than it would with a single-query ($q=1$) because the gradient estimations are executed sequentially, even though the total compute remains the same. While this trade-off improves final model accuracy, our objective is to maintain high accuracy while minimizing per-step execution time via the parallelism introduced in MobiZO.

\begin{table}[h]
\resizebox{\columnwidth}{!}{%
\begin{tabular}{ccccc}
\hline
\textbf{Query}      & \textbf{Batch size} & \textbf{Training steps} & \textbf{Performance}  & \textbf{Wall-clock time}   \\ \hline
1          & $B$        & $T$            & \ding{55}    & \ding{51} \\ \hline
$q$        & $B$        & $T/q$          & \ding{55}    & \ding{51} \\ \hline
$q$        & $B/q$      & $T$            & \ding{51}    & RGE \ding{55} $/$ MobiZO \ding{51} \\ \hline
\end{tabular}%
}
\caption{Different trade-offs for RGE.}
\label{tab:trade-off}
\end{table}

Our motivation for this adjustment stems from the convergence analysis in the MeZO work \cite{malladi2023finetuning}, which shows that reaching $\epsilon$-suboptimality requires

\small
\begin{equation*}
 t = \mathcal{O}\left( \frac{\ell}{\mu} \left( 1 + \frac{r}{n} \right) \left( 1 + \frac{\alpha}{\mu B} \right) \log\left( \frac{\mathcal{L}(\theta_0)-\mathcal{L}^*}{\epsilon} \right) \right), 
\end{equation*}
\normalsize

\noindent where $n$ is the number of zeroth-order queries, $B$ is the batch size, $r$ is the local effective rank of the Hessian, and $\alpha$ bounds the trace of the gradient-covariance matrix. Among these variables, $\ell$, $\mu$, and $\alpha$ are properties of the loss landscape that are generally not controllable, but $n$ and $B$ are hyperparameters that can be tuned. We hypothesize that increasing $n$ has a greater impact on convergence speed than increasing $B$, especially in large language models with low effective Hessian rank \cite{malladi2023finetuning}. Given the complexity of these models, analytically estimating $r$, $\alpha$, or $\mu$ is intractable; we therefore validate this hypothesis through empirical results.

To support our analysis, Table~\ref{tab:multi-q} presents averaged performance across different values of $q$ for both TinyLlama-1.1B and Llama2-7B on their corresponding tasks described in Section~\ref{sec:exp}. As $q$ increases and $E$ decreases proportionally (keeping total $B$ fixed), model performance consistently improves compared to single query RGE, confirming the benefit of our trade-off strategy. However, the improvement does not persist uniformly as $q$ continues to grow. Therefore, in practice, we recommend using $q=4$ or $q=16$, which offer a favorable balance between convergence quality and hyperparameter search.

\begin{table}[h]
\centering
\resizebox{0.8\columnwidth}{!}{%
\begin{tabular}{llllll}
\hline
$q$            & 1     & 2     & 4     & 8     & 16    \\ \hline
TinyLlama-1.1B & 70.33 & 73.62 & 74.10 & 74.82 & 74.80 \\
Llama2-7B      & 71.62 & 71.75 & 72.87 & 72.81 & 72.83 \\ \hline 
\end{tabular}%
}
\caption{Average performance of TinyLlama-1.1B and Llama2-7B models under varying numbers of queries $q$ in MebiZO.}
\label{tab:multi-q}
\end{table}

\section{Ablation Studies on System Performance of MobiZO}
\label{app:ablation}

\subsection{Efficiency of outer-loop parallelization}

We measure the runtime and memory usage of MobiZO, implemented using outer-loop parallelization only for the Llama2-7B model across different effective batch sizes and fixed sequence lengths configurations. As shown in Table \ref{tab:ablation-outer}, the runtime remains nearly identical across different combinations of the number of queries $q$ and effective batch size $E$, given that the batch size remains constant at $B=16$, which indicates our outer-loop parallelization implementation does not incur computation overhead. Peak memory usage increases slightly due to the instantiation of multiple LoRA trainable parameters at each layer.

\begin{table}[h]
\centering
\resizebox{\columnwidth}{!}{%
\begin{tabular}{llllllllll}
\hline
Sequence length      & \multicolumn{3}{c}{64} & \multicolumn{3}{c}{128} & \multicolumn{3}{c}{256} \\ \hline
$q$                  & 1      & 4     & 16    & 1      & 4      & 16    & 1      & 4      & 16    \\ \hline
Effective batch size & 16     & 4     & 1     & 16     & 4      & 1     & 16     & 4      & 1     \\ \hline
Runtime (sec/step)   & 0.18   & 0.20  & 0.19  & 0.35   & 0.37   & 0.32  & 0.69   & 0.67   & 0.71  \\ \hline
Memory (GB)          & 12.61  & 12.69 & 12.81 & 12.64  & 12.80  & 13.14 & 12.70  & 13.04  & 13.53 \\ \hline
\end{tabular}%
}
\caption{System performance of outer-loop parallelization for Llama2-7B under the same batch size of 16.}
\label{tab:ablation-outer}
\end{table}

\input{figures/app_quantization}

\subsection{Efficiency of inner-loop parallelization}

We measure the runtime and memory usage of MobiZO, implemented using inner-loop parallelization only for the Llama2-7B model across fixed different sequence length and batch size configurations. As shown in Table \ref{tab:ablation-inner}, the runtime speedup is up to $1.79\times$ at a sequence length of 64 and batch size of 1. This improvement is primarily due to reusing model weights across two forward passes, which reduces cache access and increases operation intensity. However, the benefits diminish as operation intensity increases and the system becomes compute-bound.

\begin{table}[h]
\centering
\resizebox{\columnwidth}{!}{%
\begin{tabular}{llllllllll}
\hline
Sequence length      & \multicolumn{3}{c}{64} & \multicolumn{3}{c}{128} & \multicolumn{3}{c}{256} \\ \hline
Batch size           & 1      & 8     & 16    & 1      & 8      & 16    & 1      & 8      & 16     \\ \hline
MeZO ($q=1$, LoRA-FA)& 0.07   & 0.11  & 0.18  & 0.07   & 0.19   & 0.35  & 0.07   & 0.35   & 0.69  \\ \hline
MobiZO ($q=1$, inner) & 0.04   & 0.10  & 0.18  & 0.04   & 0.18   & 0.34  & 0.06   & 0.34   & 0.67 \\ \hline
\end{tabular}%
}
\caption{Runtime (sec/step) of inner-loop parallelization for Llama2-7B under different sequence length and batch size configurations.}
\label{tab:ablation-inner}
\end{table}

Additionally, we evaluate the speedup achieved by inner-loop parallelization under weight-only INT8 and NF4 quantization. As illustrated in Figure \ref{fig:quantization}, inner-loop parallelization achieves the greatest speedup in conjunction with NF4 quantization, reaching up to a $1.97\times$ improvement over the sequential execution of two forward passes. Since NF4 dequantization is more computationally intensive than INT8 during forward passes, inner-loop parallelization enhances efficiency by dequantizing weights only once per training step, reducing the overhead from repeated dequantization.

\subsection{End-to-end training efficiency}

Tables \ref{tab:e2e-runtime1} - \ref{tab:e2e-memory2} provide additional details on per-task runtime and memory usage to complement the experimental results in Table \ref{tab:model-perf}. In these tables, MeZO (Full) represents the baseline configuration in which all model parameters are updated during training. For MeZO (LoRA-FA), results are presented for both the standard implementation without optimizations and a variant enhanced with inner-loop parallelization. For MobiZO, results are shown for two setups: one using only outer-loop parallelization and another that combines both inner and outer-loop parallelization strategies. As noted in Section \ref{sec:server-side}, when both parallelization strategies are enabled, MobiZO achieves speedups of up to $4.3\times$ over MeZO (Full) and up to $1.9\times$ over MeZO (LoRA-FA).

Regarding memory usage, enabling both inner and outer-loop parallelization results in higher memory consumption for both models compared to configurations using only outer-loop parallelization. This increase is due to the concurrent computation of two forward passes when inner-loop parallelization is enabled. Specifically, for Llama2-7B, tasks like MultiRC see an increase in memory usage of up to $33\%$ when using inner-loop parallelization due to larger sequence length. Despite this increase, the memory efficiency remains within acceptable bounds.

\section{Edge Devices Specifications}
\label{app:edge}

Table \ref{tab:edge_device} presents the specifications of the edge computing devices used in the experiments, detailing the CPU, memory, and accelerator components.

\begin{table}[h]
\centering
\resizebox{\columnwidth}{!}{
\begin{tabular}{llll}
\hline
\textbf{Device}         & \textbf{CPU}                  & \textbf{Memory}    & \textbf{Accelerator}        \\ \hline
NVIDIA Jetson           & 6$\times$ 1.5GHz Cortex-      & 8GB 68GB/s         & 1024-core Ampere            \\ 
Orin Nano               & A78AE                         & LPDDR5             & GPU 625MHz                  \\ \hline
OnePlus 12              & 1$\times$ 3.3GHz Cortex-X4    & 12GB 77GB/s        & Hexagon NPU                 \\
                        & 3$\times$ 3.2GHz Cortex-A720  & LPDDR5             &                             \\
                        & 2$\times$ 3.0GHz Cortex-A720  &                    &                             \\
                        & 2$\times$ 2.3GHz Cortex-A520  &                    &                             \\ \hline
\end{tabular}
}
\caption{Edge devices used in the experiments.}
\label{tab:edge_device}
\end{table}

For experiments on the Android NPU backend, we use the Qualcomm AI Engine Direct SDK (v2.28.0.241029) and Android NDK (r26d) to compile the kernel library.

\clearpage

\begin{table*}[t]
\centering
\resizebox{0.65\textwidth}{!}{%
\begin{tabular}{lcccccc}
\hline
\textbf{Methods} & \textbf{SST-2} & \textbf{RTE} & \textbf{MRPC} & \textbf{QQP} & \textbf{QNLI} & \textbf{WNLI} \\
\hline
MeZO (Full) ($q=1$)                 & 38.31          & 61.51        & 45.71         & 40.76        & 46.30         & 43.57         \\ \hline
MeZO (LoRA-FA) ($q=1$)              &                &              &               &              &               &               \\
\quad standard                      & 34.66          & 55.53        & 35.45         & 35.00        & 37.44         & 34.40         \\
\quad inner                         & 23.55          & 54.07        & 35.72         & 28.76        & 36.59         & 33.22         \\ \hline
MobiZO ($q=4$)                       &                &              &               &              &               &               \\
\quad outer only                    & 36.27          & 45.22        & 36.90         & 36.19        & 35.33         & 37.23         \\
\quad inner + outer                 & 23.68          & 43.75        & 34.07         & 25.83        & 31.97         & 29.09         \\ \hline
MobiZO ($q=16$)                      &                &              &               &              &               &               \\
\quad outer only                    & 35.57          & 38.18        & 35.38         & 35.19        & 35.86         & 35.34         \\
\quad inner + outer                 & 24.77          & 31.98        & 29.90         & 24.31        & 27.43         & 25.84         \\ \hline
\end{tabular}
}
\caption{Runtime (min/task) of fine-tuning TinyLlama-1.1B across different tasks using different ZO methods.}
\label{tab:e2e-runtime1}
\end{table*}

\begin{table*}[t]
\centering
\resizebox{0.9\textwidth}{!}{%
\begin{tabular}{lcccccccc}
\hline
\textbf{Methods} & \textbf{SST-2} & \textbf{RTE} & \textbf{BoolQ} & \textbf{WSC} & \textbf{WiC} & \textbf{MultiRC} & \textbf{COPA} & \textbf{WinoGrande}  \\
\hline
MeZO (Full) ($q=1$)                 & 159.44         & 288.10       & 384.07         & 209.72       & 173.01       & 526.49           & 146.40        & 154.74                       \\ \hline
MeZO (LoRA-FA) ($q=1$)              &                &              &                &              &              &                  &               &                                     \\
\quad standard                      & 54.20          & 213.81       & 329.46         & 116.79       & 70.55        & 504.74           & 40.77         & 48.07                        \\
\quad inner                         & 55.22          & 210.30       & 322.64         & 118.03       & 72.75        & 505.54           & 36.57         & 48.62                        \\ \hline
MobiZO ($q=4$)                       &                &              &                &              &              &                  &               &                                     \\
\quad outer only                    & 49.11          & 165.53       & 251.63         & 91.87        & 66.55        & 505.70           & 44.65         & 49.01                        \\
\quad inner + outer                 & 45.17          & 164.21       & 248.55         & 92.17        & 67.52        & 496.32           & 37.38         & 46.89                        \\ \hline
MobiZO ($q=16$)                      &                &              &                &              &              &                  &               &                                     \\
\quad outer only                    & 43.91          & 111.80       & 171.84         & 71.14        & 60.31        & 438.24           & 41.96         & 46.41                        \\
\quad inner + outer                 & 36.99          & 111.54       & 171.14         & 72.40        & 61.10        & 421.41           & 35.91         & 43.41                        \\ \hline
\end{tabular}
}
\caption{Runtime (min/task) of fine-tuning Llama2-7B across different tasks using different ZO methods.}
\label{tab:e2e-runtime2}
\end{table*}

\begin{table*}[t]
\centering
\resizebox{0.65\textwidth}{!}{%
\begin{tabular}{lcccccc}
\hline
\textbf{Methods} & \textbf{SST-2} & \textbf{RTE} & \textbf{MRPC} & \textbf{QQP} & \textbf{QNLI} & \textbf{WNLI} \\
\hline
MeZO (Full) ($q=1$)                 & 2.56           & 3.38         & 2.74          & 2.74         & 3.17          & 2.77          \\ \hline
MeZO (LoRA-FA) ($q=1$)              &                &              &               &              &               &               \\
\quad standard                      & 2.35           & 3.27         & 2.63          & 2.63         & 3.06          & 2.66          \\
\quad inner                         & 2.63           & 4.46         & 3.18          & 3.18         & 4.04          & 3.24          \\ \hline
MobiZO ($q=4$)                       &                &              &               &              &               &               \\
\quad outer only                    & 2.37           & 3.29         & 2.65          & 2.65         & 3.07          & 2.68          \\
\quad inner + outer                 & 2.67           & 4.50         & 3.22          & 3.22         & 4.07          & 3.28          \\ \hline
MobiZO ($q=16$)                      &                &              &               &              &               &               \\
\quad outer only                    & 2.44           & 3.18         & 2.72          & 2.69         & 3.14          & 2.75          \\
\quad inner + outer                 & 2.81           & 4.28         & 3.36          & 3.30         & 4.22          & 3.42          \\ \hline
\end{tabular}
}
\caption{Peak memory usage (GB) of fine-tuning TinyLlama-1.1B across different tasks using different ZO methods.}
\label{tab:e2e-memory1}
\end{table*}

\begin{table*}[t]
\centering
\resizebox{0.9\textwidth}{!}{%
\begin{tabular}{lcccccccc}
\hline
\textbf{Methods} & \textbf{SST-2} & \textbf{RTE} & \textbf{BoolQ} & \textbf{WSC} & \textbf{WiC} & \textbf{MultiRC} & \textbf{COPA} & \textbf{WinoGrande}  \\
\hline
MeZO (Full)                         & 13.64          & 16.23        & 18.39          & 14.51        & 13.82        & 18.39            & 13.60         & 13.60                         \\ \hline
MeZO (LoRA-FA) ($q=1$) &     &     &     &     &          \\
\quad standard                      & 13.41          & 16.00        & 18.16          & 14.27        & 13.58        & 18.16            & 12.98         & 13.15                         \\
\quad inner                         & 14.23          & 19.41        & 23.73          & 15.96        & 14.57        & 23.73            & 13.37         & 13.71                         \\ \hline
MobiZO ($q=4$) &     &     &     &     &          \\
\quad outer only                    & 13.53          & 16.12        & 18.28          & 14.40        & 13.71        & 18.28            & 13.10         & 13.27                         \\
\quad inner + outer                 & 14.47          & 19.65        & 23.97          & 16.20        & 14.82        & 23.97            & 13.61         & 13.95                         \\ \hline
MobiZO ($q=16$) &     &     &     &     &          \\
\quad outer only                    & 14.03          & 16.10        & 18.77          & 14.92        & 14.20        & 18.77            & 13.59         & 13.77                        \\
\quad inner + outer                 & 15.45          & 19.59        & 24.95          & 17.17        & 15.79        & 24.95            & 14.58         & 14.93                         \\ \hline
\end{tabular}
}
\caption{Peak memory usage (GB) of fine-tuning Llama2-7B across different tasks using different ZO methods.}
\label{tab:e2e-memory2}
\end{table*}

\end{document}

%% file: figures/1_method.tex
\begin{figure}[t]
    \centering
    \scalebox{0.75}{ 
    \begin{tikzpicture}
        \node[draw=black, fill=gray!20, text width=2.6cm, minimum height=.6cm, align=center, rounded corners](weight){\footnotesize $W \in \mathbb{R}^{k \times k}$};

        \draw[dashed, draw=black, fill=gray!5] ($(weight.west)+(-1.1,-1.5)$) rectangle ($(weight.east)+(1.0,-2.5)$);
        
        \node[below=.5cm of weight, xshift=.5cm, draw=black, fill=pink!20, text width=1.4cm, minimum height=.45cm, align=center, rounded corners](B1){\footnotesize $B_1$};
        \node[below=.5cm of B1, draw=black, fill=pink!20, text width=1.4cm, minimum height=.45cm, align=center, rounded corners](B2){\footnotesize $B_2$};
        \node[below=.2cm of B2, fill=none, text width=1.4cm, minimum height=.45cm, align=center, rounded corners, inner sep=-.1cm](Bi){\footnotesize $\vdots$};
        \node[below=.2cm of Bi, draw=black, fill=pink!20, text width=1.4cm, minimum height=.45cm, align=center, rounded corners](Bn){\footnotesize $B_q$};

        \node[left=0cm of B1, yshift=.35cm, draw=black, fill=green!20, align=center, rounded corners](z1){\tiny $z_1$};
        \node[left=0cm of B2, yshift=.35cm, draw=black, fill=green!20, align=center, rounded corners](z2){\tiny $z_2$};
        \node[left=0cm of Bn, yshift=.35cm, draw=black, fill=green!20, align=center, rounded corners](zn){\tiny $z_q$};
        \draw[->] (z1.south) to [out=-90, in=180] (B1.west);
        \draw[->] (z2.south) to [out=-90, in=180] (B2.west);
        \draw[->] (zn.south) to [out=-90, in=180] (Bn.west);

        \node[left=1.6cm of B1, draw=black, fill=brown!30, align=center, minimum height=.5cm, rounded corners](x1){\footnotesize $x_{1}$};
        \node[left=1.6cm of B2, draw=black, fill=brown!30, align=center, minimum height=.5cm, rounded corners](x2){\footnotesize $x_{2}$};
        \node[left=1.8cm of Bi, fill=none, align=center, minimum height=.5cm, rounded corners](xi){\footnotesize $\vdots$};
        \node[left=1.6cm of Bn, draw=black, fill=brown!30, align=center, minimum height=.5cm, rounded corners](xn){\footnotesize $x_q$};
        \draw[thin, ->] (x1) to (B1);
        \draw[thin, ->] (x2) to (B2);
        \draw[thin, ->] (xn) to (Bn);
        
        \node[right=.5cm of B1, draw=black, fill=brown!30, align=center, minimum height=.5cm, rounded corners](y1){\footnotesize $y_1$};
        \node[right=.5cm of B2, draw=black, fill=brown!30, align=center, minimum height=.5cm, rounded corners](y2){\footnotesize $y_2$};
        \node[right=.7cm of Bi, fill=none, align=center, minimum height=.5cm, rounded corners](yi){\footnotesize $\vdots$};
        \node[right=.5cm of Bn, draw=black, fill=brown!30, align=center, minimum height=.5cm, rounded corners](yn){\footnotesize $y_q$};
        \draw[->] (B1) to (y1);
        \draw[->] (B2) to (y2);
        \draw[->] (Bn) to (yn);

        \node[below=.5cm of weight, xshift=-1.25cm, draw=black, fill=gray!20, text width=.35cm, minimum height=2.9cm, align=center, rounded corners](A){\footnotesize $A$};
        \foreach \i in {x1,x2,xn} {
            \draw[->, rounded corners=6pt, blue!50] (\i.east) -- +(.2cm, 0cm) |- (weight.west);
        }
        \foreach \i in {y1,y2,yn} {
            \draw[-, rounded corners=6pt, blue!50] (weight.east) -- +(.2cm, 0cm) |- (\i.west);
        }

        \node[right=1cm of y2, yshift=.4cm, draw=black, fill=brown!30, align=center, minimum height=.5cm, rounded corners](xipos){\footnotesize $x_i^+$};
        \node[right=1cm of y2, yshift=-.4cm, draw=black, fill=brown!30, align=center, minimum height=.5cm, rounded corners](xineg){\footnotesize $x_i^-$};
        \node[right=.3cm of xipos, yshift=-.4cm, draw=black, fill=gray!20, text width=.35cm, minimum height=1.7cm, align=center, rounded corners](AA){\footnotesize $A$};
        \node[right=.2cm of AA, yshift=.4cm, draw=black, fill=pink!20, text width=1.4cm, minimum height=.45cm, align=center, rounded corners](Bipos){\footnotesize $B_i + z_i$};
        \node[right=.2cm of AA, yshift=-.4cm, draw=black, fill=pink!20, text width=1.4cm, minimum height=.45cm, align=center, rounded corners](Bineg){\footnotesize $B_i - z_i$};
        \node[right=.4cm of Bipos, draw=black, fill=brown!30, align=center, minimum height=.5cm, rounded corners](yipos){\footnotesize $y_i^+$};
        \node[right=.4cm of Bineg, draw=black, fill=brown!30, align=center, minimum height=.5cm, rounded corners](yineg){\footnotesize $y_i^-$};

        \draw[->] (y2.east) +(.2cm, 0cm) -- +(0.7cm, 0cm) node[xshift=-0.23cm, align=center] {\tiny for each \\ \tiny $x_i$ };

        \draw[-] (xipos.east) -- +(.3cm, 0cm);
        \draw[-] (xineg.east) -- +(.3cm, 0cm);
        \draw[-] (Bipos.west) -- +(-0.2cm, 0cm);
        \draw[-] (Bineg.west) -- +(-0.2cm, 0cm);
        \draw[->] (Bipos) to (yipos);
        \draw[->] (Bineg) to (yineg);

        \draw[decorate, decoration={brace,amplitude=4pt, mirror, raise=4pt}] 
        ($(xipos.west)+(0.1,0.4)$) -- ($(xineg.west)+(0.1,-0.4)$);

        \node[right=2.2cm of weight, draw=black, minimum width=1.6cm, fill=gray!20, align=center, rounded corners](frozen){\footnotesize Frozen};
        \node[right=0.5cm of frozen, draw=black, minimum width=1.6cm, fill=pink!20, align=center, rounded corners](trainable){\footnotesize Trainable};

        \node[below=0.3cm of Bn, draw=none, fill=none, xshift=-.3cm](part1){(a) Outer-loop};
        \node[right=3.0cm of part1, draw=none, fill=none](part2){(b) Inner-loop};
    \end{tikzpicture}
    }
    \caption{Overview of the MP-LoRA module in the MobiZO framework that supports both outer-loop and inner-loop parallelization, enabling faster training and improved model accuracy with minimal memory usage and access overhead.}
    \label{fig:method}
\end{figure}

%% file: figures/3_executorch.tex
\begin{figure*}[t]
    \centering

    \resizebox{\textwidth}{!}{ 
    \begin{tikzpicture}[node distance=2cm, every node/.style={scale=0.8}, thick]
    
        \tikzstyle{block1} = [rectangle, rounded corners, minimum width=2cm, minimum height=1cm, text centered, draw=black, fill=gray!20, align=center]
        \tikzstyle{block2} = [rectangle, rounded corners, minimum width=2cm, minimum height=1cm, text centered, draw=black, fill=green!20, align=center]
        \tikzstyle{block3} = [rectangle, minimum width=1cm, minimum height=0.3cm, text centered, draw=black, fill=green!20, align=center]
        \tikzstyle{line} = [draw, ->]
    
        \node (model) [block1] { Model \\ Authoring};
        \node (lora) [block2, right of=model, xshift=1.7cm] { Mobile MP-\\ LoRA module};
        \node (graph) [block1, right of=lora, xshift=1.7cm] { Exported \\ Graph};
        \node (program) [block1, right of=graph, xshift=1.5cm] { ExecuTorch \\ Program};
        \node (load) [block1, right of=program, xshift=2.2cm] { Load Data \\ \& Program};
        \node (execute) [block1, right of=load, xshift=1cm] { Execute};
        \node (kernel) [block1, right of=execute, xshift=1cm] { Kernel Library \\};
        \node (op) [block3, below of=kernel, yshift=1.71cm] {\tiny Custom Op};
    
        \node (pic1) [left of=model, xshift=0.5cm] {\includegraphics[width=1cm]{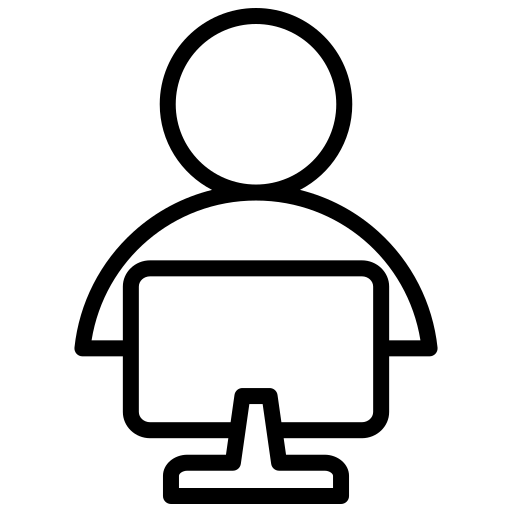}};
        \node (pic4) [right of=kernel, xshift=-0.3cm] {\includegraphics[width=1cm]{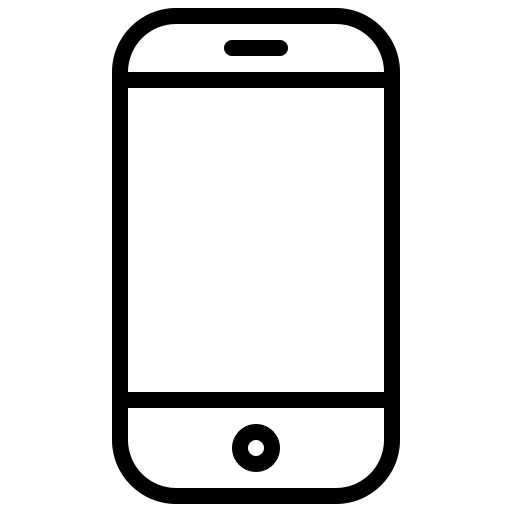}};
        
        \draw [line] (model) -- node[above] {modify} node[below] {\tiny \texttt{nn.Module}} (lora);
        \draw [line] (lora) -- node[above] {export} node[below] {\tiny \texttt{nn.Module}} (graph); 
        \draw [line] (graph) -- node[above] {compile} node[below] {\tiny \texttt{ExportIR}} (program);
        \draw [line] (program) -- node[above] {offload} node[below] {\tiny \texttt{flatbuffer}} (load);
        \draw [line, ->, bend left] (load) to (execute);
        \draw [line, ->, bend left] (execute) to (load);
        \draw [line] (kernel) -- (execute);

        Dashed boxes
        \draw[dashed, thick] ($(model.north west)+(-0.8,0.3)$) rectangle ($(program.south east)+(0.2,-0.3)$);
        \draw[dashed, thick] ($(load.north west)+(-0.2,0.3)$) rectangle ($(kernel.south east)+(0.8,-0.3)$);
    
        \node at ($(lora.north) + (1, 0.55)$) {Offline Compile-time (Server)};
        \node at ($(execute.north)+ (0.3, 0.55)$) {Online Runtime (Edge)};
    
    \end{tikzpicture}
    } 

\caption{MobiZO on-device training workflow via ExecuTorch with minimal modifications. The green box represents additional procedure in addition to standard steps for inference deployment on edge devices.}
\label{fig:executorch}
\end{figure*}

%% file: figures/4_end2end_time.tex
\pgfplotstableread{
Task MeZO MeZO_LoRA_FA P_RGE_q4 P_RGE_q16
SST2 38.31 34.66 23.68 24.77
RTE 61.51 55.53 43.75 31.98
MRPC 45.71 35.45 34.07 29.90
QQP 40.76 35.00 25.83 24.31
QNLI 46.30 37.44 31.97 27.43
WNLI 43.57 34.40 29.09 25.84
}\datatableone


\pgfplotstableread{
Task MeZO MeZO_LoRA_FA P_RGE_q4 P_RGE_q16
SST2 159.44 54.20 45.17 36.99
RTE 288.10 213.81 164.21 111.54
BoolQ 384.07 329.46 248.55 171.14
WSC 209.72 116.79 92.17 72.40
WiC 173.01 70.55 67.52 61.10
MultiRC 526.49 504.74 496.32 421.41
COPA 146.40 40.77 37.38 35.91
WinoGrande 154.74 48.07 46.89 43.41
}\datatabletwo

\begin{figure*}[t]
    \small
    \hspace{10pt}
    \begin{subfigure}[b]{0.43\textwidth}
        \centering
        \begin{tikzpicture}[trim axis left, trim axis right]
            \begin{axis}[
                width=\textwidth,
                height=4cm,
                title={TinyLlama-1.1B},
                title style={yshift=-4pt},
                ybar=1pt,
                ymin=0,
                ymax=79,
                enlarge x limits=0.15,
                bar width=4pt,
                ylabel={Runtime (min/task)},
                ylabel style={yshift=-10pt},
                symbolic x coords={SST2, RTE, MRPC, QQP, QNLI, WNLI},
                xtick=data,
                xtick pos=lower,
                xticklabel style={
                    rotate=30,
                    anchor=east,
                    anchor=north east,
                    align=right,
                    font=\tiny,
                    text width=1.5em
                },
                ymajorgrids=true,
                grid style={dashed, gray!30},
                nodes near coords,
                every node near coord/.append style={
                    font=\tiny,
                    xshift=5pt,
                    yshift=7pt,
                    rotate=90,
                    /pgf/number format/.cd,
                    fixed,
                    fixed zerofill,
                    precision=0
                },
            ]
            \addplot table [x=Task, y=MeZO] {\datatableone};
            \addplot table [x=Task, y=MeZO_LoRA_FA] {\datatableone};
            \addplot table [x=Task, y=P_RGE_q4] {\datatableone};
            \addplot table [x=Task, y=P_RGE_q16] {\datatableone};

            \end{axis}
        \end{tikzpicture}
    \end{subfigure}%
    \hspace{-20pt}
    \small
    \begin{subfigure}[b]{0.5\textwidth}
        \centering
        \begin{tikzpicture}[trim axis left, trim axis right]
            \begin{axis}[
                width=\textwidth,
                height=4cm,
                title={Llama2-7B},
                title style={yshift=-4pt},
                ybar=1pt,
                ymin=0,
                ymax=700,
                enlarge x limits=0.07,
                bar width=4pt,
                symbolic x coords={SST2, RTE, BoolQ, WSC, WiC, MultiRC, COPA, WinoGrande, SQuAD},
                xtick=data,
                xtick pos=lower,
                xticklabel style={
                    rotate=30,
                    anchor=north east,
                    align=right,
                    font=\tiny,
                    text width=1.5em
                },
                ymajorgrids=true,
                grid style={dashed, gray!30},
                nodes near coords,
                every node near coord/.append style={
                    font=\tiny,
                    xshift=5pt,
                    yshift=7pt,
                    rotate=90,
                    /pgf/number format/.cd,
                    fixed,
                    fixed zerofill,
                    precision=0
                },
                legend style={at={(1.0,0.5)}, anchor=west, font=\tiny},
                legend columns=1,             
                legend image code/.code={\draw[#1, draw=none] (0cm,-0.1cm) rectangle (0.2cm,0.1cm); },
            ]
            \addplot table [x=Task, y=MeZO] {\datatabletwo};
            \addplot table [x=Task, y=MeZO_LoRA_FA] {\datatabletwo};
            \addplot table [x=Task, y=P_RGE_q4] {\datatabletwo};
            \addplot table [x=Task, y=P_RGE_q16] {\datatabletwo};
            \legend{MeZO (Full), MeZO (LoRA-FA), MobiZO ($q=4$), MobiZO ($q=16$)};
            \end{axis}
        \end{tikzpicture}
    \end{subfigure}
    
    \caption{End-to-end wall-clock time of fine-tuning TinyLlama-1.1B and Llama2-7B for various configurations across tasks. MobiZO achieves up to 4.3$\times$ speedup compared to MeZO (full) under the same computational budget.}
    \label{fig:end2end-time}
\end{figure*}

%% file: figures/5_NPU.tex
\pgfplotstableread[row sep=\\,col sep=&]{
batch & runtime \\
b2 & 0.37 \\
b4 & 0.63 \\
b8 & 1.24 \\
b16 & 2.80 \\
}\androidAruntime

\pgfplotstableread[row sep=\\,col sep=&]{
batch & runtime \\
b2 & 0.93 \\
b4 & 1.67 \\
b8 & 2.95 \\
b16 & 5.76 \\
}\androidBruntime

\pgfplotstableread[row sep=\\,col sep=&]{
batch & memory \\
b2 & 3.37 \\
b4 & 3.54 \\
b8 & 3.77 \\
b16 & 3.88 \\
}\androidAmemory

\pgfplotstableread[row sep=\\,col sep=&]{
batch & memory \\
b2 & 3.44 \\
b4 & 3.70 \\
b8 & 3.93 \\
b16 & 4.49 \\
}\androidBmemory

\begin{figure}[t]
\centering
\begin{tabular}{cc}
    \begin{tikzpicture}
    \begin{axis}[
        ybar=0pt,
        bar width=5pt,
        width=4.3cm,
        height=3.5cm,
        enlarge x limits=0.2,
        xlabel={Batch size},
        xlabel style={yshift=4pt},
        ylabel={Runtime (sec/step)},
        ylabel style={yshift=-20pt},
        symbolic x coords={b2, b4, b8, b16},
        xtick=data,
        ymin=0,
        ymax=8,
        font=\scriptsize,
        tick label style={font=\tiny},
        xticklabels={2, 4, 8, 16},
        xticklabel style={yshift=2pt},
        xtick pos=lower,
        ymajorgrids=true,
        grid style={dashed, gray!30},
        nodes near coords,
        every node near coord/.append style={
            font=\tiny,
            xshift=5pt,
            yshift=9pt,
            rotate=90,
            /pgf/number format/.cd,
        },
        group style={
            group bars=true,
            group width=12pt,
            horizontal sep=6pt
        },
        legend style={at={(0.05,1)}, anchor=north west, font=\tiny, draw=none, fill=none, nodes={scale=0.8, transform shape}},
        legend image post style={scale=0.5}, 
        legend cell align={left},
        legend image code/.code={\draw[#1, draw=none] (0cm,-0.1cm) rectangle (0.3cm,0.1cm);},
    ]
    \addplot table[x=batch, y=runtime] {\androidAruntime};
    \addplot table[x=batch, y=runtime] {\androidBruntime};
    \legend{$Seq Len=64$, $Seq Len=128$};
    \end{axis}
    \end{tikzpicture}
    &
    \begin{tikzpicture}
    \begin{axis}[
        ybar=0pt,
        bar width=5pt,
        width=4.3cm,
        height=3.5cm,
        enlarge x limits=0.2,
        xlabel={Batch size},
        xlabel style={yshift=4pt},
        ylabel={Memory (GB)},
        ylabel style={yshift=-20pt},
        symbolic x coords={b2, b4, b8, b16},
        xtick=data,
        ymin=2,
        ymax=6,
        font=\scriptsize,
        tick label style={font=\tiny},
        xticklabels={2, 4, 8, 16},
        xticklabel style={yshift=2pt},
        xtick pos=lower,
        ymajorgrids=true,
        grid style={dashed, gray!30},
        nodes near coords,
        every node near coord/.append style={
            font=\tiny,
            xshift=5pt,
            yshift=9pt,
            rotate=90,
            /pgf/number format/.cd,
        },
        group style={
            group bars=true,
            group width=12pt,
            horizontal sep=6pt
        }
    ]
    \addplot table[x=batch, y=memory] {\androidAmemory};
    \addplot table[x=batch, y=memory] {\androidBmemory};
    \end{axis}
    \end{tikzpicture}
\end{tabular}
\caption{Runtime and memory usage per step on Android NPU backend for TinyLlama-1.1B across different batch sizes and sequence lengths.}
\label{fig:NPU}
\end{figure}

%% file: figures/2_padding.tex
\begin{figure}[h]
    \centering
    \small
    \begin{subfigure}[b]{0.46\columnwidth}
        \centering
        \begin{tikzpicture}
            \begin{axis}[
                xbar stacked,
                width=4cm, height=4cm,
                title={Padding on batch size of 8},
                xlabel={Token Count},
                ylabel={Sample Index},
                ytick={1, 2, 3, 4, 5, 6, 7, 8},
                yticklabels={1, 2, 3, 4, 5, 6, 7, 8},
                y dir=reverse,
                xmin=0, xmax=128,
                bar width=0.28cm,
                enlarge y limits=0.1,
                enlarge x limits=0.05,
                xtick={1, 32, 64, 96, 128},
                yticklabel style={anchor=east},
                ylabel style={yshift=-15pt}, 
            ]
        
            \addplot coordinates {(40,1) (60,2) (100,3) (50,4) (30,5) (128,6) (70,7) (90,8)};
            \addplot coordinates {(88,1) (68,2) (28,3) (78,4) (98,5) (0,6) (58,7) (38,8)};

            \end{axis}
        \end{tikzpicture}
    \end{subfigure}
    \hfill
    \begin{subfigure}[b]{0.46\columnwidth}
        \centering
        \begin{tikzpicture}
            \begin{axis}[
                xbar stacked,
                width=4cm, height=4cm,
                title={Padding on batch size of 2},
                xlabel={Token Count},
                ytick={1, 2, 3, 4, 5, 6, 7, 8},
                yticklabels={1, 2, 3, 4, 5, 6, 7, 8},
                y dir=reverse,
                xmin=0, xmax=128,
                bar width=0.28cm,
                enlarge y limits=0.1,
                enlarge x limits=0.05,
                xtick={1, 32, 64, 96, 128},
                yticklabel style={anchor=east}
            ]
        
            \addplot coordinates {(40,1) (60,2) (100,3) (50,4) (30,5) (128,6) (70,7) (90,8)};
            \addplot coordinates {(20,1) (0,2) (0,3) (50,4) (98,5) (0,6) (20,7) (0,8)};
        
            \end{axis}
        \end{tikzpicture}
    \end{subfigure}

    \caption{The standard batching approach pads shorter sequences to the maximum sequence length within the batch.}
    \label{fig:padding}
\end{figure}

%% file: figures/app_padding_stats.tex
\pgfplotstableread{
Task        Batch16    Batch4
sst2        57.83      38.27
rte         51.76      32.69
boolq       50.37      32.24
wsc         44.58      24.11
wic         24.76      13.66
multirc     3.94       1.31
copa        25.88      16.39
winogrande  22.79      10.42
mrpc        24.90      16.07
qqp         35.22      19.08
qnli        36.21      21.75
wnli        36.28      20.56
}\datatable

\begin{figure}[h]
    \centering
    \begin{tikzpicture}
        \begin{axis}[
            width=\columnwidth,
            height=5cm,
            ybar=0.1,
            ymin=0, ymax=60,
            bar width=4pt,
            ylabel={\small Padding Tokens Avg Percentage},
            ylabel style={yshift=-10pt},
            symbolic x coords={sst2, rte, boolq, wsc, wic, multirc, copa, winogrande, mrpc, qqp, qnli, wnli},
            xtick=data,
            xtick pos=lower,
            xticklabel style={
                rotate=45,
                anchor=east,
                font=\small
            },
            ymajorgrids=true,
            grid style={dashed, gray!30},
            legend entries={\small Batch size 16, \small Batch size 4},
            legend image code/.code={
                \draw[#1, draw=none] (0cm,-0.1cm) rectangle (0.3cm,0.1cm);
            },
        ]
        \addplot table [x=Task, y=Batch16] {\datatable};
        \addplot table [x=Task, y=Batch4] {\datatable};
        \end{axis}
    \end{tikzpicture}
    \vspace{-2pt}
    \caption{Average percentage of padding tokens for different tasks and batch sizes.}
    \label{fig:padding_tokens}
\end{figure}

%% file: figures/app_quantization.tex
\begin{figure*}[t]
\centering
\resizebox{\textwidth}{!}{
\begin{tabular}{cccccc}
    \multicolumn{3}{c}{TinyLlama-1.1B} & \multicolumn{3}{c}{Llama2-7B} \\  
    \begin{tikzpicture}
        \begin{axis}[
            title={seq len = 64},
            xlabel={batch size},
            ylabel={Runtime speedup},
            ylabel style={yshift=-5pt},
            grid=major,
            width=3.8cm,
            height=5cm,
            xmin=0, xmax=17,
            ymin=1, ymax=2,
            font=\small,
            xtick={1, 8, 16},
        ]
        \addplot[color=green, mark=triangle, style=solid] coordinates {
            (1,1.728) (8,1.732) (16,1.692)
        };
        \addplot[color=red, mark=square, style=solid] coordinates {
            (1,1.802) (8,1.675) (16,1.654)
        };
        \addplot[color=blue, mark=o, style=solid] coordinates {
            (1,1.508) (8,1.399) (16,1.375)
        };
        \end{axis}
    \end{tikzpicture}
    &
    \begin{tikzpicture}
        \begin{axis}[
            title={seq len = 128},
            xlabel={batch size},
            grid=major,
            width=3.8cm,
            height=5cm,
            xmin=0, xmax=17,
            ymin=1, ymax=2,
            font=\small,
            xtick={1, 8, 16},
        ]
        \addplot[color=green, mark=triangle, style=solid] coordinates {
            (1,1.845) (8,1.673) (16,1.433)
        };
        \addplot[color=red, mark=square, style=solid] coordinates {
            (1,1.653) (8,1.559) (16,1.211)
        };
        \addplot[color=blue, mark=o, style=solid] coordinates {
            (1,1.532) (8,1.432) (16,1.000)
        };
        \end{axis}
    \end{tikzpicture}
    &
    \begin{tikzpicture}
        \begin{axis}[
            title={seq len = 256},
            xlabel={batch size},
            grid=major,
            width=3.8cm,
            height=5cm,
            xmin=0, xmax=17,
            ymin=1, ymax=2,
            font=\small,
            xtick={1, 8, 16},
            legend pos=north east,
            legend style={font=\small, draw=none, fill=none,nodes={scale=0.8, transform shape}},
        ]
        \addplot[color=green, mark=triangle, style=solid] coordinates {
            (1,1.821) (8,1.475) (16,1.257)
        };
        \addplot[color=red, mark=square, style=solid] coordinates {
            (1,1.694) (8,1.204) (16,1.097)
        };
        \addplot[color=blue, mark=o, style=solid] coordinates {
            (1,1.591) (8,1.050) (16,1.056)
        };
        \legend{NF4, INT8, FP16} 
        \end{axis}
    \end{tikzpicture}
    &
    \begin{tikzpicture}
        \begin{axis}[
            title={seq len = 64},
            xlabel={batch size},
            ylabel={Runtime speedup},
            ylabel style={yshift=-5pt},
            grid=major,
            width=3.8cm,
            height=5cm,
            xmin=0, xmax=17,
            ymin=1, ymax=2,
            font=\small,
            xtick={1, 8, 16},
        ]
        \addplot[color=green, mark=triangle, style=solid] coordinates {
            (1,1.970) (8,1.303) (16,1.291)
        };
        \addplot[color=red, mark=square, style=solid] coordinates {
            (1,1.820) (8,1.371) (16,1.176)
        };
        \addplot[color=blue, mark=o, style=solid] coordinates {
            (1,1.788) (8,1.112) (16,1.032)
        };
        \end{axis}
    \end{tikzpicture}
    &
    \begin{tikzpicture}
        \begin{axis}[
            title={seq len = 128},
            xlabel={batch size},
            grid=major,
            width=3.8cm,
            height=5cm,
            xmin=0, xmax=17,
            ymin=1, ymax=2,
            font=\small,
            xtick={1, 8, 16},
        ]
        \addplot[color=green, mark=triangle, style=solid] coordinates {
            (1,1.948) (8,1.463) (16,1.332)
        };
        \addplot[color=red, mark=square, style=solid] coordinates {
            (1,1.758) (8,1.176) (16,1.107)
        };
        \addplot[color=blue, mark=o, style=solid] coordinates {
            (1,1.673) (8,1.043) (16,1.024)
        };
        \end{axis}
    \end{tikzpicture}
    &
    \begin{tikzpicture}
        \begin{axis}[
            title={seq len = 256},
            xlabel={batch size},
            grid=major,
            width=3.8cm,
            height=5cm,
            xmin=0, xmax=17,
            ymin=1, ymax=2,
            font=\small,
            xtick={1, 8, 16},
            legend pos=north east,
            legend style={font=\small, draw=none, fill=none,nodes={scale=0.8, transform shape}},
        ]
        \addplot[color=green, mark=triangle, style=solid] coordinates {
            (1,1.914) (8,1.180) (16,1.116)
        };
        \addplot[color=red, mark=square, style=solid] coordinates {
            (1,1.666) (8,1.106) (16,1.078)
        };
        \addplot[color=blue, mark=o, style=solid] coordinates {
            (1,1.302) (8,1.025) (16,1.036)
        };
        \legend{NF4, INT8, FP16} 
        \end{axis}
    \end{tikzpicture}
\end{tabular}
}
\caption{Runtime speedup per training step of TinyLlama-1.1B and Llama2-7B for different quantization methods, sequence lengths, and batch sizes.}
\label{fig:quantization}
\end{figure*}